\documentclass[manuscript]{acmart}
\AtBeginDocument{%
  }

\setcopyright{none}
\renewcommand\footnotetextcopyrightpermission[1]{}
\renewcommand\footnotetextauthorsaddresses[1]{}

\usepackage{booktabs}
\usepackage{multirow}
\usepackage{tabularx}
\usepackage{colortbl}
\usepackage[normalem]{ulem}
\usepackage{graphicx}
\usepackage{pifont}

\definecolor{aliceblue}{rgb}{0.94, 0.97, 1.0}

\hypersetup{
  pdftitle={SignGPT: Toward LLM-Mediated Sign Language Interaction through Gloss-Free Translation and Generation},
  pdfauthor={Ronghui Li, Jun Dong, Zhongyuan Hu, Zunnan Xu, Jun Zhou, Liyuan Chen, Shuoling Liu, Jiangpeng Yan, Jie Guo, Xiu Li, and Linchao Bao},
  pdflang={en-US}
}

\begin{document}

\fancypagestyle{firstpagestyle}{%
  \fancyhf{}
  \fancyfoot[C]{\thepage}
  \renewcommand{\headrulewidth}{0pt}
  \renewcommand{\footrulewidth}{0pt}}
\fancypagestyle{arxivpagestyle}{%
  \fancyhf{}
  \fancyfoot[C]{\thepage}
  \renewcommand{\headrulewidth}{0pt}
  \renewcommand{\footrulewidth}{0pt}}

\title{SignGPT: Toward LLM-Mediated Sign Language Interaction through Gloss-Free Translation and Generation}

\author{Ronghui Li}
\authornote{Co-first authors.}
\affiliation{%
  \institution{Tsinghua University}
  \city{Beijing}
  \country{China}}

\author{Jun Dong}
\authornotemark[1]
\affiliation{%
  \institution{Tsinghua University}
  \city{Shenzhen}
  \country{China}}

\author{Zhongyuan Hu}
\affiliation{%
  \institution{Tsinghua University}
  \city{Shenzhen}
  \country{China}}

\author{Zunnan Xu}
\affiliation{%
  \institution{Nanyang Technological University}
  \city{Singapore}
  \country{Singapore}}

\author{Jun Zhou}
\affiliation{%
  \institution{Tsinghua University}
  \city{Shenzhen}
  \country{China}}

\author{Liyuan Chen}
\affiliation{%
  \institution{E Fund}
  \city{Guangzhou}
  \country{China}}

\author{Shuoling Liu}
\affiliation{%
  \institution{E Fund}
  \city{Guangzhou}
  \country{China}}
\email{liushuoling@efunds.com.cn}

\author{Jiangpeng Yan}
\affiliation{%
  \institution{E Fund}
  \city{Guangzhou}
  \country{China}}

\author{Jie Guo}
\affiliation{%
  \institution{Peng Cheng Laboratory}
  \city{Shenzhen}
  \country{China}}

\author{Xiu Li}
\authornote{Corresponding author: \href{mailto:li.xiu@sz.tsinghua.edu.cn}{li.xiu@sz.tsinghua.edu.cn}.}
\affiliation{%
  \institution{Tsinghua University}
  \city{Shenzhen}
  \country{China}}
\email{li.xiu@sz.tsinghua.edu.cn}

\author{Linchao Bao}
\affiliation{%
  \institution{Tencent AI Lab}
  \city{Shenzhen}
  \state{Guangdong}
  \country{China}}
\email{linchaobao@tencent.com}

\renewcommand{\shortauthors}{Li et al.}

\begin{abstract}
Large language models (LLMs) provide limited support for sign language interaction. Unifying sign language translation (SLT) and generation (SLG) to enable sign language as both input and output can reduce switching between separate models during sign–-text interaction. We present SignGPT, a unified, pose-based framework for gloss-free SLT and SLG. SignGPT integrates part-aware hierarchical representations of body, hand, and facial motion into a shared language model and employs asymmetric multi-token prediction and progressive training for bidirectional modeling. We evaluate SignGPT on How2Sign (ASL) and Phoenix-2014T (DGS) through benchmark comparisons, qualitative analyses, and component ablations. An exploratory study with 12 Deaf ASL signers assesses an LLM-mediated sign-to-sign response pipeline, highlighting the potential of unified modeling to support sign language conversation (SLC). See the project page: \url{https://signgpt-demo.github.io/sign-language-interaction-demo/}.

\end{abstract}



\begin{teaserfigure}
  \includegraphics[width=\textwidth]{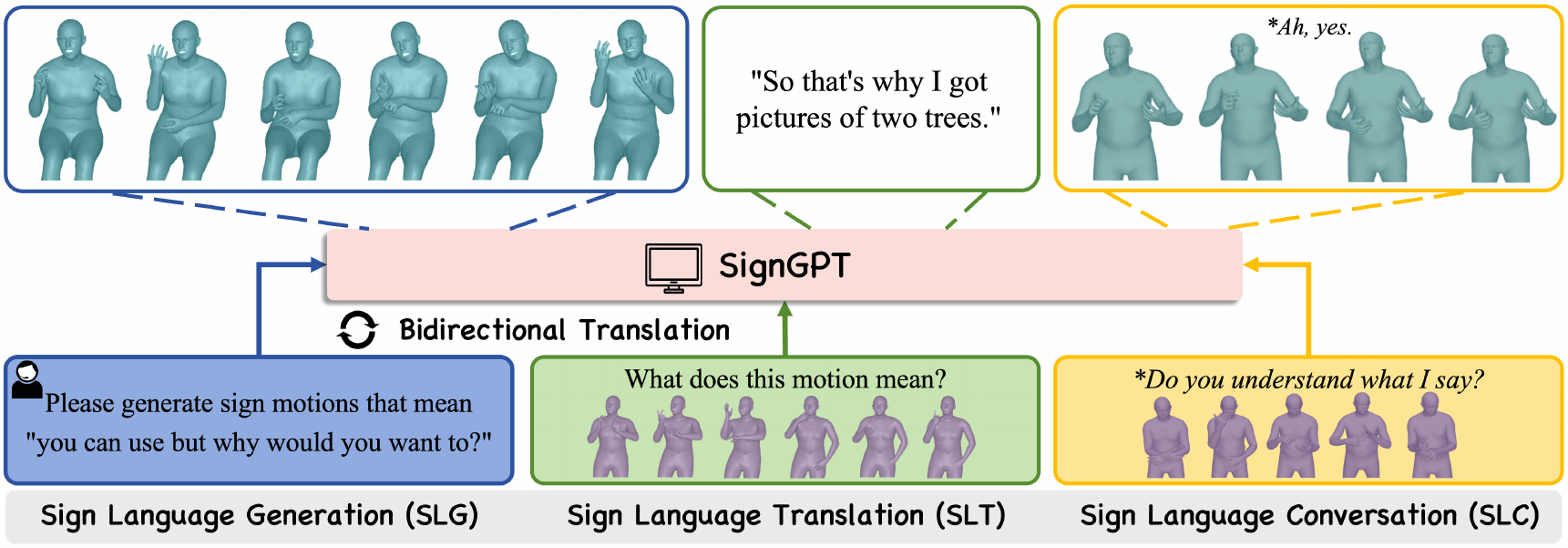}
  \caption{A central SignGPT model connects translation and generation. On the left, a text instruction is converted into a sequence of signing avatars. In the middle, an input signing sequence is translated into text. On the right, a signed question is translated, answered in English, and converted into a signed response, illustrating the one-turn response pipeline.}
  \Description{SignGPT jointly models sign-to-text translation and text-to-sign generation within each dataset-specific model, and we explore how these complementary capabilities can be composed in a one-turn sign-to-sign response pipeline.}
  \label{fig:teaser}
\end{teaserfigure}

\maketitle
\pagestyle{arxivpagestyle}
\thispagestyle{arxivpagestyle}

\section{Introduction}
Sign languages are full-fledged natural languages, each with its own lexicon, grammar, and regional variation~\cite{500}. Yet mainstream artificial intelligence interfaces are designed primarily around speech and written text. This creates interaction barriers for people who prefer to communicate in sign language and often requires them to repeatedly switch between signing and written text. Recent human--computer interaction (HCI) research further argues that sign language technologies must account for non-manual markers, user agency, and translation practices within Deaf communities, rather than reducing sign language to generic body motion~\cite{zhang2025nonmanual,tang2026reimagining}.

Bidirectional sign--text systems could reduce such modality switching by supporting sign language as both input and output. However, a useful sign-to-sign response pipeline must preserve linguistically meaningful handshape, orientation, location, movement, and non-manual information in both directions. Most existing work studies sign language translation (SLT) and sign language generation (SLG) separately, making it difficult to examine how errors propagate across sign language understanding, text-mediated response generation, and motion synthesis. Recent gloss-free and multilingual approaches have improved performance in individual mapping directions~\cite{hwang2025efficient,tan2025multilingual,lai2026selective}, while large-scale benchmarks continue to reveal substantial gaps in the sign language understanding capabilities of current multimodal models~\cite{zhao2026cnslbench}.

Learning both sign-to-text and text-to-sign mappings within a single model exposes two related representational challenges. First, motion tokenizers designed for general human motion may inadequately capture the fine-grained hand articulation and non-manual markers that convey lexical, grammatical, and pragmatic meaning. Second, token-based language models typically append discrete motion codes to their vocabularies with newly initialized embeddings, making these codes difficult to align with the pretrained model's textual semantic representations. These problems can become further compounded within a sign language interaction pipeline. Because SLT and SLG are typically modeled and evaluated in isolation, downstream failures caused by errors in input representation or translation are difficult to diagnose. Sign-to-sign interaction therefore calls for a shared framework that connects motion representation, bidirectional language mapping, and pipeline-level evaluation.

To this end, we present SignGPT, a unified pose-based framework that supports both SLT and SLG (Figure~\ref{fig:teaser}). Its Part-aware Hierarchical VQ-VAE (PHVQ) combines body-to-hand hierarchical quantization with bidirectional multiscale temporal encoding to represent coordinated body, hand, and facial motion. Its Gloss-free Heterogeneous Motion-aware Language Model (GHMLM) reuses PHVQ features as motion embeddings, mitigating the mismatch between learned motion representations and language-model inputs. Through Asymmetric Multi-Token Prediction (AMTP), GHMLM predicts either text tokens or part-specific motion tokens from shared hidden states, thereby supporting gloss-free sign-to-text translation and text-to-sign generation. We further connect these capabilities in an exploratory LLM-mediated sign-to-sign response pipeline.

We evaluate SignGPT on How2Sign~\cite{143} and Phoenix-2014T~\cite{141}, which cover American Sign Language (ASL) and German Sign Language (DGS), respectively. Our quantitative and qualitative analyses include component ablations and address three research questions. \textbf{RQ1 (Motion Representation):} How well does part-aware hierarchical tokenization preserve coordinated body, hand, and facial motion? \textbf{RQ2 (Unified Modeling):} How does SignGPT perform on gloss-free SLT and SLG under established ASL and DGS benchmark protocols, and how do its key components affect performance? \textbf{RQ3 (Response Experience):} How do raters assess the appropriateness and motion smoothness of responses produced by the LLM-mediated pipeline, and what limitations do these ratings reveal?

Our contributions are threefold:

\begin{itemize}
    \item We present SignGPT, a unified pose-based framework that connects gloss-free sign-to-text translation and text-to-sign generation through part-aware motion quantization, shared hidden states, and heterogeneous prediction heads.

    \item We provide benchmark comparisons and component ablations on ASL and DGS datasets, separately evaluating motion reconstruction quality and task-level translation and generation performance.

    \item We construct an exploratory LLM-mediated sign-to-sign response pipeline and examine raters' perceptions of response appropriateness and motion smoothness, characterizing the opportunities and current limitations of LLM-mediated sign language interaction.
\end{itemize}

\section{Related Work}
\subsection{Sign Language Translation (SLT)}
SLT maps visual--spatial linguistic input into written text, requiring models to capture manual and non-manual signals distributed across space and time. Conventional systems typically adopt a gloss-mediated pipeline: a sign language recognition model (often trained with CTC) first predicts a gloss sequence from RGB video, and a language model then translates the glosses into text~\cite{slt,107}. Although glosses provide structured intermediate supervision, annotating them is costly and requires language-specific expertise, which has motivated a shift toward direct video-to-text translation through temporal modeling~\cite{502}, pretrained visual and language representations~\cite{503,112,signllm-slt}, contextual information~\cite{sltcc}, and lexical or semantic supervision~\cite{505}.

Recent gloss-free methods focus on interfacing signed input with pretrained language models: explicitly modeling spatial configurations and motion dynamics~\cite{hwang2025efficient}, aligning sign representations with language-model representations~\cite{inan2025signalignlm}, applying selective contrastive learning~\cite{lai2026selective}, or scaling training to multiple sign languages~\cite{tan2025multilingual}. Complementary evaluations expose fingerspelling and cross-lingual transfer weaknesses that aggregate metrics obscure~\cite{tanzer2025fingerspelling,tanzer2025fleurs}; three-dimensional ASL modeling offers a directly comparable reference point for pose-based translation~\cite{zhang2026largesign}, and CNSL-bench documents persistent sign language understanding gaps in multimodal language models~\cite{zhao2026cnslbench}.

These efforts are directed primarily at the sign-to-text direction. SignGPT instead employs PHVQ pose representations within a shared architecture and, through Asymmetric Multi-Token Prediction, supports both gloss-free sign-to-text translation and text-to-sign generation.

\subsection{Sign Language Generation (SLG)}

SLG maps linguistic input to temporally coordinated body, hand, and facial motion. Recent methods include diffusion-based generation~\cite{nsa,115} and autoregressive motion modeling with language-model architectures~\cite{116,momask}. SOKE discretizes signing into part-specific token sequences and combines autoregressive generation with retrieval~\cite{soke}; SIGNLLM targets multilingual sign production~\cite{506}; other approaches construct long sequences by composing retrieved segments~\cite{120} or smoothing transitions between gloss-conditioned segments~\cite{121}. Recent work further addresses transition-pose generation~\cite{tang2025transition}, semantics-aware evaluation~\cite{imai2025silverscore}, and personalized continuous production~\cite{rastgoo2026personalized}.

Linguistically adequate signing, however, involves more than smooth body motion---non-manual markers themselves carry lexical, grammatical, and pragmatic information~\cite{zhang2025nonmanual}---yet most SLG systems still optimize production independently of sign understanding. Building on part-specific motion modeling, SignGPT combines body-to-hand hierarchical quantization with multiscale temporal encoding and reuses the resulting pose representations across SLG and SLT.

\subsection{Language Models for Tokenized Motion}

Research on general human-motion generation laid the groundwork for discretizing continuous pose sequences into tokens~\cite{124}: T2M-GPT couples a VQ-VAE with GPT-based autoregressive synthesis~\cite{t2mgpt}, MotionGPT treats motion tokens as a foreign language to jointly model text and motion~\cite{motiongpt}, and subsequent work extends this formulation to motion understanding and instruction following~\cite{127,128,motiongpt2}. These methods show that text and motion can be handled within a unified sequence-modeling framework, but general-purpose tokenizers often fail to allocate sufficient capacity to the fine-grained hand articulation and non-manual signals that sign languages require, and motion codes are typically merely appended to the vocabulary with embeddings learned from scratch during downstream training. SignGPT maps PHVQ's learned motion-code representations into motion-token embeddings via learned projectors, providing a structured, motion-informed interface while leaving motion--language alignment to be learned during training.

\subsection{Human-Centered Sign Language Interaction}

Human-centered research frames sign language translation and generation as situated practices in which meaning is continually created, interpreted, and negotiated. ELMI shows how automatic assistance can be integrated into interactive authoring for song signing~\cite{yoo2025elmi}; studies with Deaf online creators indicate that translation involves audience expectations, identity, platform constraints, and repeated repair rather than a one-shot transfer between linguistic forms~\cite{tang2026reimagining}; and research with ASL educators reveals both the opportunities and the risks of AI-supported learning, underscoring the importance of context and user agency~\cite{hassan2026asl}. These findings remind us that kinematic smoothness is not sufficient evidence of communicative success. We therefore report automatic motion metrics separately from human ratings, and position our LLM-mediated response pipeline as an exploratory probe of system-level output quality.

Overall, prior work has advanced gloss-free SLT, SLG, and token-based motion modeling separately, but these components are largely developed and evaluated in isolation. SignGPT combines part-aware pose tokenization with shared sign--text modeling, reuses learned motion-code representations, and explores a sign-to-sign response pipeline.

\begin{figure*}[t]
\centering
\includegraphics[width=\textwidth]{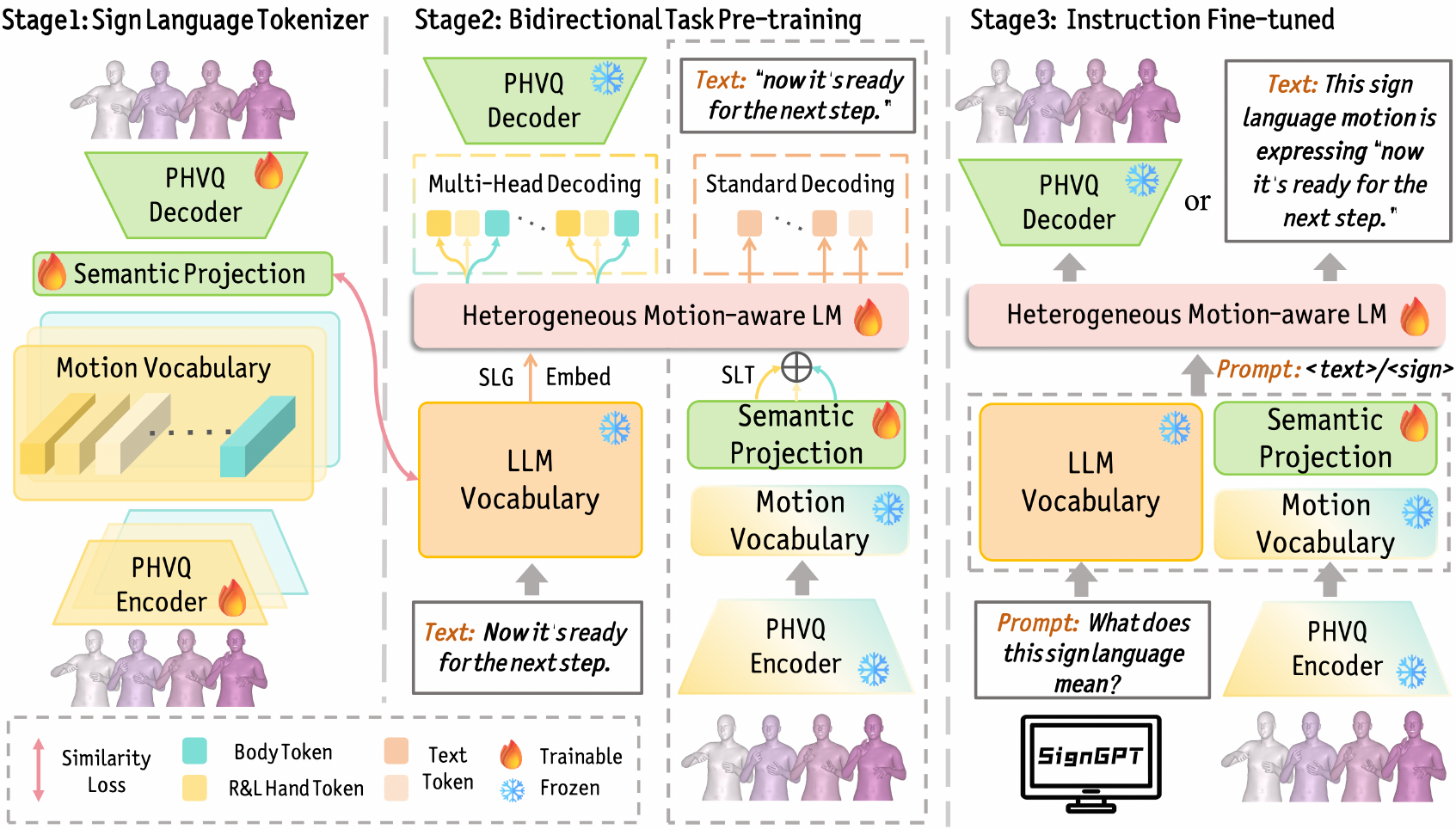}
\caption{Overview of SignGPT's three training stages: (1) PHVQ discretizes continuous body and hand motion, with an optional text-alignment objective; (2) GHMLM is jointly trained on SLG and SLT and uses AMTP to decode text or part-aware motion tokens; and (3) instruction fine-tuning supports prompted translation and generation.}
\label{fig:overview}
\Description{A three-stage pipeline. Stage 1 trains PHVQ encoders, codebooks, a projector--reconstructor path, and a decoder; an optional frozen text-embedding branch supplies the alignment target. Stage 2 freezes the tokenizer and trains a heterogeneous motion-aware language model on text-to-sign and sign-to-text directions, using separate motion heads and a standard text head. Stage 3 freezes selected components and instruction-tunes text and motion prompts for translation and generation use. Flame and snowflake icons distinguish trainable and frozen modules.}
\end{figure*}

\section{Methodology}
We present SignGPT, a unified pose-based framework for sign language translation (SLT) and sign language generation (SLG). SignGPT represents signing as discrete, part-aware motion sequences and couples these sequences with text in a pretrained language model. As shown in Fig.~\ref{fig:overview}, the framework contains two main components. (1) The Part-aware Hierarchical VQ-VAE (PHVQ) encodes full-body signing into compact discrete representations. Its body-to-hand hierarchical quantization preserves global postural context while modeling fine-grained hand articulation, and an optional objective regularizes the quantized motion features toward the paired textual feature space. (2) The Gloss-free Heterogeneous Motion-aware Language Model (GHMLM) retrieves quantized PHVQ features as motion embeddings and maps them into the LLM hidden space using learned projectors. It fuses embeddings across body parts and predicts text or part-specific motion tokens from shared hidden states through heterogeneous output heads. We train these components progressively using PHVQ pretraining, joint LoRA-based optimization of SLT and SLG, and instruction fine-tuning. Subsequently, we combine the trained SignGPT with an external language model as an intermediary to construct an exploratory sign-to-sign response pipeline.

\subsection{Motion Representation}
We use SMPL-X-based 3D motion to obtain a compact representation of signing while reducing visual redundancy. The native SMPL-X parameterization uses local joint rotations, which do not explicitly encode relative spatial relationships between articulators (e.g., the proximity of an index finger to the nose). Inspired by HumanML3D~\cite{h3d}, we instead use joint positions to represent these spatial relationships directly. Following prior work~\cite{humantomato,151}, we omit per-joint velocity and rotation features while retaining root motion. Specifically, a motion sequence is represented as $\mathcal{M}\in\mathbb{R}^{N\times230}$, where $N$ is the number of frames. Each frame is $m_i=(\dot{r}_a,\dot{r}_x,\dot{r}_z,r_y,\mathbf{j}_p,\mathbf{f})$: a one-dimensional root-yaw angular velocity $\dot{r}_a$, two-dimensional root linear velocity $(\dot{r}_x,\dot{r}_z)$ on the ground plane, scalar root height $r_y$, root-relative 3D coordinates $\mathbf{j}_p$ for 72 non-root joints, and 10 SMPL-X facial-expression parameters $\mathbf{f}$. The dimensionality is therefore $1+2+1+(72\times3)+10=230$.

\subsection{Sign Language Tokenization}
\begin{figure*}[t]
\centering
\includegraphics[width=\textwidth]{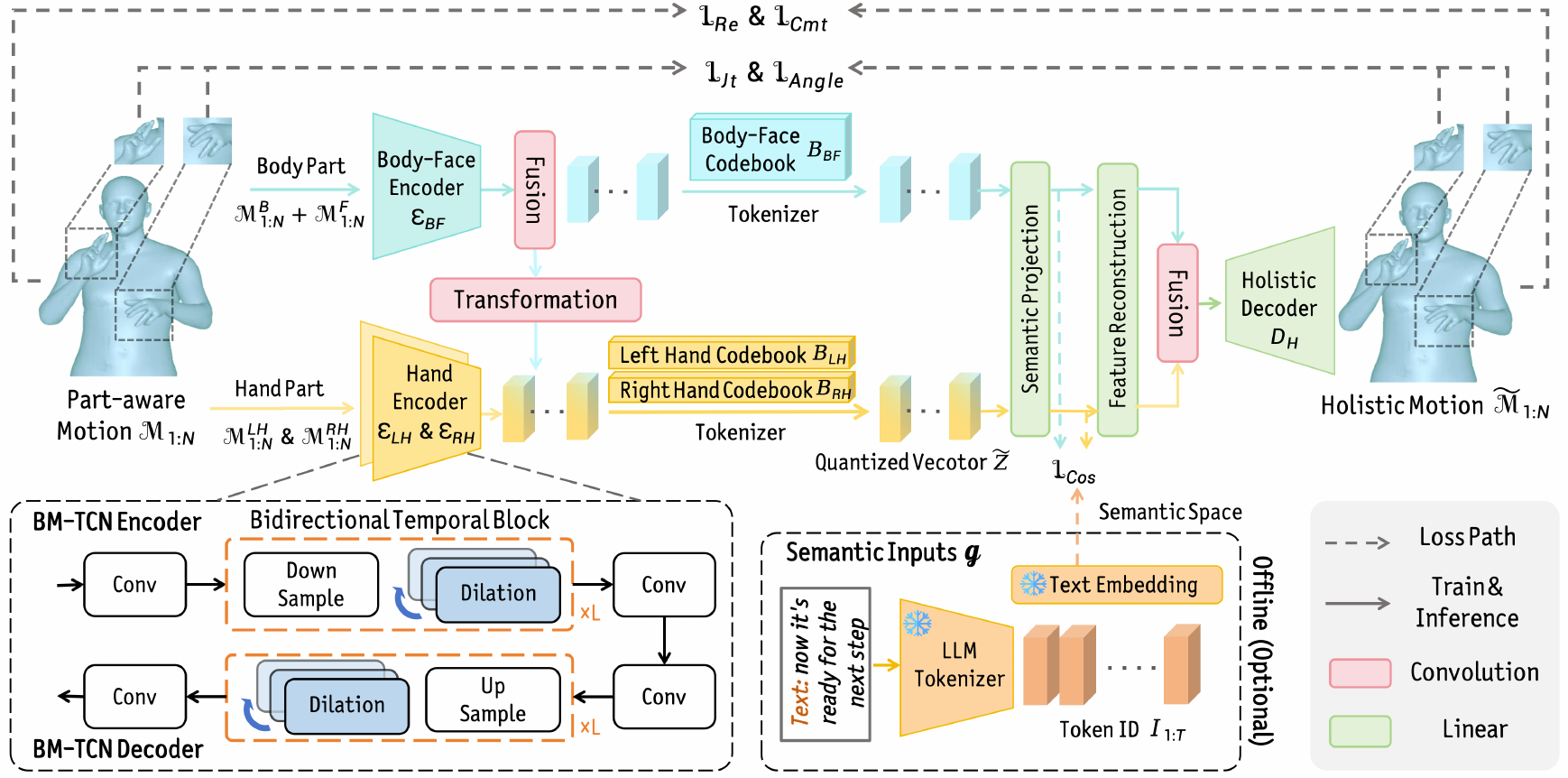}
\caption{Overview of the Part-aware Hierarchical VQ-VAE (PHVQ) tokenizer. PHVQ combines bidirectional multi-scale temporal modeling with body-to-hand hierarchical quantization and encodes a full-body sign language motion sequence into three discrete streams: a body--face codebook $B_{Body}$, a left-hand codebook $B_{LH}$, and a right-hand codebook $B_{RH}$. Facial-expression features are encoded jointly with the body stream.}
\label{fig:phvq}
\Description{A flow diagram separates an input motion into body--face, left-hand, and right-hand features. A body--face encoder and codebook produce global context, which is transformed and passed to the left- and right-hand encoders before their separate codebooks. The three quantized streams are projected, fused, and decoded into a holistic motion sequence. Insets show the bidirectional multi-scale temporal encoder--decoder and the optional frozen text-embedding branch used for cosine alignment. Dashed arrows denote loss paths, and solid arrows denote the training and inference flow.}
\end{figure*}

Residual vector quantization (RVQ)~\cite{508} successively quantizes errors left by preceding codebook layers and can improve motion reconstruction~\cite{509}. We do not use residual codebook stacks in PHVQ. Combining $L$ residual levels with three part streams would require $3L$ discrete targets at each motion timestep; serializing these targets would lengthen the autoregressive sequence, whereas predicting them jointly would enlarge the synchronized output space. Either choice would substantially increase language-model optimization complexity. PHVQ therefore uses one vector-quantization lookup for each of the body--face, left-hand, and right-hand streams. ``Hierarchical'' refers to body-to-hand conditioning. Building on prior motion tokenizers~\cite{motiongpt2,humantomato}, PHVQ introduces three design elements: (i) body-to-hand hierarchical quantization, which conditions the hand streams on quantized body-level context; (ii) a Bidirectional Multi-scale Temporal Convolutional Network (BM-TCN) that encodes past and future context over multiple temporal scales; and (iii) an optional motion--text alignment objective that encourages the quantized representations to approach the paired textual feature space. Fig.~\ref{fig:phvq} shows the PHVQ architecture.

\subparagraph{\textbf{Part-aware Motion Decomposition and Body-to-Hand Hierarchical Quantization.}}
Given a full-body sign motion sequence $\mathcal{M}=[m_1,\ldots,m_N]$ of $N$ frames, where $m_i\in\mathbb{R}^{230}$, we decompose each frame into body, left-hand, right-hand, and facial-expression components. We concatenate the body and facial features as $\mathcal{M}_\text{BF}=[\mathcal{M}_\text{Body};\mathcal{M}_\text{Face}]$ and process them jointly. A dedicated temporal encoder $\mathcal{E}^T_i$, implemented with BM-TCN, encodes each stream:
\begin{equation}
\small
Z_i = \mathcal{E}^T_i(\mathcal{M}_i), \quad i \in \{\text{BF}, \text{LH}, \text{RH}\}.
\end{equation}
Here, $Z_{\text{BF}}\in\mathbb{R}^{d_{bf}\times T'}$ and $Z_{\text{LH}},Z_{\text{RH}}\in\mathbb{R}^{d_h\times T'}$, where $d_{bf}$ and $d_h$ are the code-vector dimensions for the body--face and hand streams, respectively, and $T'=N/l$ is the temporally downsampled sequence length at downsampling rate $l$.

Body and hand motion differ in amplitude and frequency but remain anatomically and linguistically coordinated: hand articulation is contextualized by arm trajectories and torso configuration~\cite{507}. HumanTOMATO~\cite{humantomato} injects hand features into body quantization to improve whole-body coordination. We reverse this conditioning direction for signing. PHVQ first quantizes the body--face stream to obtain global postural context and then injects the resulting quantized features into the left- and right-hand quantization streams. This design is intended to support hand reconstruction with information about the accompanying arms and torso; we evaluate its contribution through ablation. A unified decoder $\mathcal{D}_T$ reconstructs the full motion sequence from the fused quantized features of all three streams. Appendix~\ref{app:phvq-quantization} provides the implementation details.

\subparagraph{\textbf{Bidirectional Multi-scale Temporal Encoding.}}
PHVQ tokenizes complete motion sequences offline, allowing each encoded position to use both preceding and subsequent context. This context is relevant to signing because anticipatory coarticulation may begin before a sign's main articulation, while transitional motion may continue afterward. Plain 1D ResNet encoders used by existing motion tokenizers~\cite{t2mgpt} rely on stacked fixed-kernel convolutions with a uniform progression of receptive fields. This design can make it difficult to represent both short finger transitions and longer phrase-level motion patterns within a compact encoder. Conventional causal TCNs~\cite{131}, developed for streaming or autoregressive settings, use only past context and therefore cannot exploit future frames during offline tokenization.

We use symmetric padding so that each temporal position can access context on both sides. The resulting Bidirectional Temporal Block (BTB) is defined as
\begin{equation}
\small
\mathrm{BTB}(x) = \mathrm{ReLU}\!\Big(\mathcal{F}^{(2)}\!\big(\mathcal{F}^{(1)}(x)\big) + \mathcal{W}_{\mathrm{res}}(x)\Big),
\end{equation}
where each BTB contains two weight-normalized 1D convolutions $\mathcal{F}^{(j)}$ with symmetric padding $p=d\cdot\lfloor(k{-}1)/2\rfloor$; $k$ is the kernel size and $d$ is the dilation factor. Each convolution is followed by a ReLU activation and dropout, and $\mathcal{W}_{\mathrm{res}}$ denotes a residual projection used when channel dimensions differ. We stack multiple BTBs with exponentially increasing dilation rates $d_i=r^i$. The resulting receptive field covers local articulation and longer temporal dependencies without requiring global attention.

\subparagraph{\textbf{Optional Text-Space Alignment.}}
We optionally regularize the quantized motion representations using paired textual features so that the resulting motion embeddings are more suitable for subsequent language modeling. For each stream $p\in\{\mathrm{BF},\mathrm{LH},\mathrm{RH}\}$, a part-specific projector $P^{\uparrow}_p$ maps the quantized latent representation into a common $d_s$-dimensional space, where $d_s=d_{\mathrm{LLM}}=2048$. The text target is produced without a separate text encoder: we use the frozen input-embedding layer of the same LLaMA 3.2-1B backbone employed by GHMLM, tokenize each paired sentence in its original corpus language (German for Phoenix-2014T and English for How2Sign), and mean-pool the valid non-padding token embeddings. The cosine loss $\mathcal{L}_{\mathrm{Cos}}$ averages the three part-specific distances between the pooled projected motion stream and the same pooled text embedding. A feature reconstructor $P^{\downarrow}_p$ maps each projected representation back to its codebook space, after which $\mathcal{D}_T$ reconstructs the motion sequence. The projector--reconstructor path is used in both variants and is trained by the reconstruction objective; SignGPT+TSA additionally enables the cosine term, whereas the base SignGPT model sets its weight to zero. We describe this objective as an embedding-alignment regularizer rather than assuming that it establishes semantic equivalence between text and signing.

\subparagraph{\textbf{Finger Representation Enhancement and Training Objective.}}
Fine-grained finger articulation carries linguistically relevant distinctions in sign language. We therefore add a hand-joint position loss $\mathcal{L}_{\text{Pos}}$ and an angle loss $\mathcal{L}_{\text{Angle}}$ to the PHVQ objective. The complete training objective is
\begin{align}
\small
\mathcal{L}_{\text {Total }} &= \mathcal{L}_{\text {Recon }} + \mathcal{L}_{\text {Pos }} + \lambda_1 \mathcal{L}_{\mathrm{Cmt}} + \lambda_2 \mathcal{L}_{\text {Cos}} \nonumber \\
&\quad + \lambda_3 \mathcal{L}_{\text {Angle }}
\label{eq:total_loss}
\end{align}
where $\mathcal{L}_{\text{Recon}}$ is the $L_2$ reconstruction error computed on the motion-representation features~\cite{t2mgpt}. The hand-specific term $\mathcal{L}_{\text{Pos}}$ is computed on the 3D hand-joint coordinates recovered from the reconstructed representation and provides direct geometric supervision for finger articulation. The implemented vector-quantization commitment loss is $\mathcal{L}_{\text{Cmt}}=\sum_i\|Z_i-\mathrm{sg}[\hat{Z}_i]\|_2^2$, where $Z_i$ is the encoder output, $\hat{Z}_i$ is its corresponding quantized codebook vector, and $\mathrm{sg}$ denotes stop-gradient. Thus, this term updates the encoder toward the selected codebook vector while blocking its direct gradient to the codebook. We instead update the codebooks using exponential moving averages (EMA)~\cite{motiongpt} and reset underused entries to reduce codebook collapse. The coefficients $\lambda_1$--$\lambda_3$ balance the objectives. Appendix~\ref{app:phvq-loss} provides formal definitions of $\mathcal{L}_{\text{Angle}}$ and $\mathcal{L}_{\text{Cos}}$.

\subsection{Gloss-free Heterogeneous Motion-aware Language Model}

\subparagraph{\textbf{Feature-informed Sign Motion Embeddings.}}
Previous motion generation and understanding methods~\cite{soke,sltcc} commonly quantize a motion sequence $\mathcal{M}$ into discrete token indices $I=\{i_1,i_2,\ldots,i_K\}$ and append these indices to the language model vocabulary. The corresponding embeddings are then initialized and learned separately from the motion tokenizer. This approach does not preserve the geometry of the tokenizer's feature space and may weaken semantic alignment between the motion and language representations.

Inspired by~\cite{138}, we retrieve the quantized feature $\hat{z}_k^p=\mathcal{C}^p[i_k^p]$ from the frozen codebook $\mathcal{C}^p$ for each stream $p\in\{\mathrm{BF},\mathrm{LH},\mathrm{RH}\}$. We initialize a trainable GHMLM projector $\widetilde{P}^{\uparrow}_p$ from the corresponding PHVQ projector and construct the motion embedding as
$e_k^p=\widetilde{P}^{\uparrow}_p(\hat{z}_k^p)\in\mathbb{R}^{d_{\mathrm{LLM}}}$.
Thus, the LLM receives an embedding derived from the learned PHVQ code vector rather than an embedding determined only by the token index. This preserves the code-to-latent correspondence at initialization; subsequent motion--language alignment is learned during GHMLM training. For SLG, text embeddings condition the autoregressive generation of fused motion embeddings. For SLT, the fused motion sequence conditions standard autoregressive text decoding.

\subparagraph{\textbf{Asymmetric Multi-Token Prediction for SLT and SLG.}}
GHMLM combines embedding-level fusion with Asymmetric Multi-Token Prediction (AMTP) to support SLT and SLG within one architecture. Here, multi-token prediction denotes the simultaneous prediction of three synchronized part tokens at one timestep, rather than the prediction of multiple future timesteps. During SLG, the three part embeddings from the preceding step are fused into one LLM input:
\begin{equation}\label{eq:fusion}\small
  \mathcal{F}(y_{k-1};\beta)
  = (1 - 2\beta)\, e_{k-1}^{\mathrm{BF}}
  + \beta\, e_{k-1}^{\mathrm{LH}}
  + \beta\, e_{k-1}^{\mathrm{RH}},
\end{equation}
where $\beta\in[0,0.5]$ is a hyperparameter controlling the relative contribution of the hand embeddings. One backbone evaluation produces a shared hidden state $h_k=f_{\mathrm{LM}}(X,\mathcal{F}(y_{k-1};\beta),\mathrm{KV}_{<k})$, where $X$ contains the instruction and textual conditioning context. Four output heads operate on this state: the original text head $\theta_{\mathrm{text}}$ and three motion heads $\{\theta_{\mathrm{BF}},\theta_{\mathrm{LH}},\theta_{\mathrm{RH}}\}$, each projecting $h_k$ into its own augmented codebook vocabulary. Sharing $h_k$ provides a common contextual signal, while separate output spaces provide part-specific supervision. The SLG distribution factorizes as
\begin{equation}\label{eq:joint}\small
  P(Y^{\mathrm{BF}},Y^{\mathrm{LH}},Y^{\mathrm{RH}}\mid X)
  =\prod_{k=1}^{K}\prod_{p\in\{\mathrm{BF},\mathrm{LH},\mathrm{RH}\}}
  P\!\left(y_k^p\mid X,Y_{<k}^{\mathrm{BF}},Y_{<k}^{\mathrm{LH}},Y_{<k}^{\mathrm{RH}}\right).
\end{equation}
The three tokens at timestep $k$ are conditionally independent given the shared state, but each head conditions on the fused history of all three streams. During SLT, the fused motion sequence forms the conditioning prefix and the text head performs standard autoregressive decoding:
\begin{equation}\label{eq:slt_generation}\small
P(T\mid\mathcal{M})=\prod_{j=1}^{|T|}P(t_j\mid\mathcal{M},t_{<j}).
\end{equation}

The instruction and an input task indicator select the decoding direction. During SLG, the three motion heads predict part-specific motion tokens, while $\theta_{\mathrm{text}}$ receives a constant auxiliary mode target. During SLT, $\theta_{\mathrm{text}}$ predicts text tokens, while the motion heads receive their respective auxiliary mode targets. We implement this behavior using dual-track labels with independent output spaces: text targets come from the LLM vocabulary, whereas BF, LH, and RH targets come from their corresponding augmented codebook vocabularies. Each inactive head receives a mode-indicator target defined within its own output space. The resulting objective is
\begin{equation}\label{eq:dual_label}\small
  \mathcal{L}_{\mathrm{total}} =
  \underbrace{\mathcal{L}_{\mathrm{CE}}^{\mathrm{text}}}_{\text{text head}}
  +\;
  \underbrace{
    \mathcal{L}_{\mathrm{CE}}^{\mathrm{BF}}
  + \mathcal{L}_{\mathrm{CE}}^{\mathrm{LH}}
  + \mathcal{L}_{\mathrm{CE}}^{\mathrm{RH}}
  }_{\text{motion heads}}.
\end{equation}

\subparagraph{\textbf{Comparison with Existing Decoding Strategies.}}
MotionGPT~\cite{motiongpt} uses one language head to decode text and coarse whole-body motion tokens from an expanded shared vocabulary. This design provides a simple bidirectional interface but does not explicitly separate body-part output spaces. SOKE~\cite{soke} predicts body-part tokens with parallel heads from a weighted-average input; its heads share a vocabulary and projection matrix and use post-hoc $-\infty$ logit masks to separate token subsets. Its encoder--decoder formulation is designed for unidirectional generation. GHMLM instead uses one LLM backbone pass per step and lightweight prediction heads with separate output dimensions and label spaces. This comparison concerns decoding structure; the empirical contribution of AMTP is evaluated in Section~\ref{sec:ablation}.

\subsection{Training and Inference Strategy}

\subparagraph{\textbf{Stage 1: Sign Language Tokenizer Training.}}
We first train PHVQ using the objective in Equation~\ref{eq:total_loss}. This stage optimizes the granularity of the quantized features and the reconstruction of full-body signing. PHVQ maps each motion sequence $\mathcal{M}$ to part-specific token-index sequences $[I_{\text{BF}},I_{\text{LH}},I_{\text{RH}}]$ and their quantized features $[\hat{Z}_{\text{BF}},\hat{Z}_{\text{LH}},\hat{Z}_{\text{RH}}]$. In subsequent stages, we freeze the PHVQ encoders, codebooks, projectors $P^{\uparrow}_p$, reconstructors $P^{\downarrow}_p$, fusion projection, and decoder $\mathcal{D}_T$. We use $\mathcal{R}_T$ to denote this complete frozen reconstruction path from part-specific token indices through codebook lookup, projection, fusion, and $\mathcal{D}_T$. We copy the PHVQ $P^{\uparrow}_p$ weights to initialize the separate, trainable GHMLM motion projectors described above.

\subparagraph{\textbf{Stage 2: Joint SLT--SLG Adaptation.}}
For each dataset-specific instance, we next optimize SLT and SLG jointly on paired sign language--text data~\cite{116,soke}, using the contextual representations of a pretrained LLM~\cite{512}. Each mini-batch contains equal numbers of SLT and SLG examples, balancing task sampling across the two directions. We apply supervised LoRA fine-tuning while freezing the original vocabulary embeddings and base model parameters. The motion projectors remain trainable so that PHVQ features can be mapped into the LLM hidden space.

\subparagraph{\textbf{Stage 3: Instruction Fine-Tuning.}}
Finally, we use task-specific instruction templates to express SLT and SLG through a common prompting interface. An SLG template is: \textit{Please generate a sign language motion sequence that expresses: ``The weather is great today.''} An SLT template is: \textit{Please accurately describe the semantic content expressed by \texttt{<sign\_tokens>}.} Here, \texttt{<sign\_tokens>} is a placeholder for the part-specific motion sequence; in implementation, we replace it with the corresponding projected embeddings derived from $[\hat{Z}_{\text{BF}},\hat{Z}_{\text{LH}},\hat{Z}_{\text{RH}}]$.

\subparagraph{\textbf{SLT and SLG Inference.}}
For both tasks, GHMLM performs autoregressive inference conditioned on the task instruction and input task indicator, using greedy decoding and one backbone evaluation per decoding step. For SLT, decoding stops when the text head $\theta_{\mathrm{text}}$ predicts an end-of-sequence (EOS) token. For SLG, the three motion streams are decoded synchronously until the implementation's joint stopping criterion is met or the maximum motion length is reached. The frozen PHVQ reconstruction path $\mathcal{R}_T$ retrieves the predicted BF, LH, and RH token indices from their respective codebooks, applies the projector--reconstructor and fusion modules, and decodes the resulting features into a continuous full-body motion sequence.

\subsection{Exploratory LLM-Mediated Sign-to-Sign Response Pipeline}

In this study, we operationalize sign language conversation (SLC) as an exploratory, single-turn sign-to-sign response task. The pipeline combines the two trained directions of SignGPT with a frozen external LLM that generates an English response. We treat it as a cascaded, LLM-mediated pipeline rather than a direct or end-to-end sign-to-sign model.

Let $V_{\mathrm{in}}$ denote an input signing video and $q$ denote the external video-to-pose preprocessing pipeline. We first obtain the pose representation $\mathcal{M}_{\mathrm{in}}=q(V_{\mathrm{in}})$ and then compute
\begin{equation}
\hat{x}=f_{\mathrm{SLT}}(\mathcal{M}_{\mathrm{in}}),\qquad
\hat{r}=g_{\mathrm{LLM}}(P,\hat{x}),\qquad
\hat{I}_{\mathrm{out}}=f_{\mathrm{SLG}}(\hat{r}),\qquad
\hat{\mathcal{M}}_{\mathrm{out}}
=\mathcal{R}_T(\hat{I}_{\mathrm{out}}),
\label{eq:slc_pipeline}
\end{equation}
where $\hat{x}$ is the English translation of the input signing, $P$ is a fixed response prompt, $\hat{r}$ is the English response generated by the external model $g_{\mathrm{LLM}}$, and $\hat{I}_{\mathrm{out}}$ contains the predicted BF, LH, and RH motion-token streams.

Throughout the response-pipeline evaluation, $g_{\mathrm{LLM}}$ is the frozen Meta Llama 3.2 1B Instruct model~\cite{512}, with the same checkpoint used in both pipelines. The fixed system prompt $P$ asks for exactly one concise response sentence in an everyday conversational style, using no more than 30 words. Only the mediator uses stochastic sampling: temperature $0.7$, top-$p$ $0.9$, top-$k$ $30$, and at most 50 new tokens. These sampling settings apply only to the mediator and do not modify the sign-model decoding procedures.

For demonstrations starting from raw video, $q$ first detects 2D body, hand, and facial landmarks using OpenPose~\cite{openpose} and lifts them to 3D using 3DposeEstimator~\cite{3dpose}. Where required, the lifted joints are fitted to the SMPL-X skeleton through inverse kinematics or optimization. We then convert the fitted sequence into the 230-dimensional representation described above, using the same spatial and temporal normalization applied to the PHVQ training data.

\section{Experiments}

\subsection{Experimental Setup}

\subparagraph{\textbf{Datasets and Evaluation Metrics.}}
We evaluate SignGPT on Phoenix-2014T~\cite{141} and How2Sign~\cite{143}. For Phoenix-2014T, we use the SMPL-X pose annotations released with SOKE~\cite{soke}; for How2Sign, we use those provided by prior work~\cite{116}.

For SLT, we report BLEU-4~\cite{148} and ROUGE-L~\cite{149}. For SLG, we report dynamic time warping with joint-position error (DTW-JPE)~\cite{146}. Because back-translation scores depend strongly on the evaluator, we do not use them as a primary SLG metric. In the one-turn response analysis, we retain cycle-consistency BLEU only as a diagnostic proxy.

To evaluate PHVQ reconstruction, we follow prior work~\cite{999} and report mean per-joint position error (MPJPE), Procrustes-aligned MPJPE (PA-MPJPE), and acceleration error (ACCEL). We report the spatial errors in millimeters.

For the exploratory one-turn response pipeline, we report stage-wise text metrics, a GPT-4o relevance rate, and human ratings of response appropriateness and motion smoothness. Section~\ref{sec:response-eval} defines these measures and their scope.

\subparagraph{\textbf{Implementation Details.}}
For PHVQ, we use codebook sizes of $N_{BF}=128$ for the body--face stream and $N_{LH}=N_{RH}=256$ for the two hand streams, with a code-vector dimension of 1024. The temporal encoders use a downsampling rate of 4. GHMLM uses the decoder-only LLaMA 3.2-1B model~\cite{512} as its language-model backbone. We apply LoRA with rank 128 and scaling parameter $\alpha=128$. All models are optimized with AdamW. PHVQ is trained for 1000 epochs with a learning rate of $10^{-4}$ and a batch size of 512. Joint SLT--SLG adaptation runs for 300 epochs with a learning rate of $2\times10^{-4}$, followed by 100 epochs of instruction fine-tuning with a learning rate of $10^{-4}$; both stages use a batch size of 64. During SLG decoding, an EOS token predicted by any motion head terminates all motion streams synchronously.

\subparagraph{\textbf{Matched MotionGPT Adaptation.}}
We adapt and retrain MotionGPT for sign language rather than transferring its published general-motion results. For each dataset, MotionGPT uses the same 230-dimensional pose sequences, paired text, preprocessing, and training, validation, and test splits as SignGPT. Its language-model backbone is likewise LLaMA 3.2-1B, and its single-stream VQ-VAE uses a 512-entry codebook. We selected this capacity in our preliminary MotionGPT tokenizer comparison. We match the training budgets and batch sizes used for SignGPT: 1000 epochs with batch size 512 for tokenizer training, 300 epochs with batch size 64 for joint SLT--SLG adaptation, and 100 epochs with batch size 64 for instruction fine-tuning. Separate MotionGPT checkpoints are trained for How2Sign and Phoenix-2014T. All MotionGPT results in this paper are reproduced using this adaptation; all remaining non-SignGPT baseline values in Tables~\ref{tab:comparison-slg} and~\ref{tab:comparison-slt} are transcribed from the source studies cited for the corresponding results.

\subsection{Comparison with Prior Methods}
\subparagraph{\textbf{Sign Language Generation.}}
Table~\ref{tab:comparison-slg} compares SignGPT with prior SLG methods. Relative to the adapted MotionGPT baseline, SignGPT+TSA reduces Avg-DTW-JPE from 9.45 to 4.32 on Phoenix-2014T and from 9.82 to 4.76 on How2Sign. Among the listed methods, SignGPT+TSA obtains the lowest hand DTW-JPE on Phoenix-2014T and the lowest body and aggregate DTW-JPE on How2Sign; SOKE retains the lowest Phoenix-2014T body and aggregate DTW-JPE and the lowest How2Sign hand DTW-JPE. MotionGPT uses a unified whole-body codebook, whereas SignGPT uses part-aware codebooks and codebook-derived language-model embeddings. The observed differences are consistent with the intended benefit of preserving part-specific information, although this cross-method comparison does not isolate any single design choice. The optional TSA variant yields small additional gains over SignGPT on the reported SLG metrics.

\begin{figure*}[t]
\centering
\includegraphics[width=\textwidth]{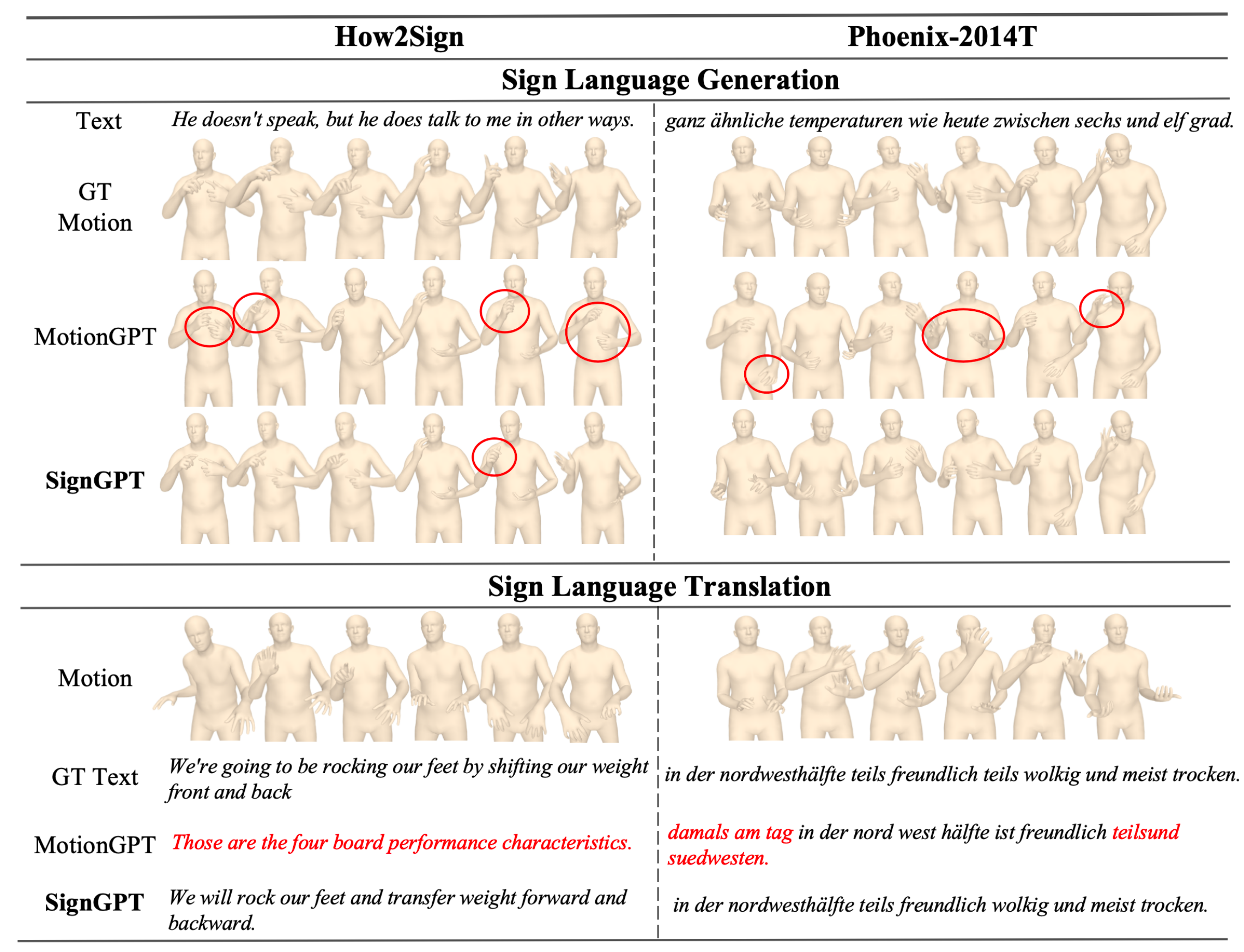}
\caption{Qualitative comparison between SignGPT and MotionGPT on SLT and SLG. The upper half shows reference, MotionGPT, and SignGPT avatar frames for text-to-sign generation on How2Sign and Phoenix-2014T; red circles mark visibly incorrect hand or arm configurations. The lower half shows input signing frames followed by reference text and the translations from both models. SignGPT outputs remain closer to the reference motion and text than MotionGPT outputs.}
\label{fig:slt-slg}
\Description{A two-dataset qualitative comparison.}
\end{figure*}

In the selected examples in Fig.~\ref{fig:slt-slg}, SignGPT exhibits fewer visible spatial artifacts and unnatural joint configurations than the adapted MotionGPT outputs. These examples are illustrative rather than a population-level comparison.

\begin{table*}[t]
    \caption{
    \textbf{Sign language generation (SLG) on Phoenix-2014T and How2Sign.} DTW-JPE is reported for the body, hands, and the aggregate labeled ``Avg'' in each implementation. Bi-T indicates support for both sign-to-text and text-to-sign directions within one dataset-specific model. SignGPT+TSA uses the optional text-embedding alignment objective with paired sentence-level translations during PHVQ training. MotionGPT results are reproduced using our matched sign-language adaptation; all other baseline values are transcribed from the source studies cited for the corresponding results. Bold and underline mark the best and second-best values.
    }
    \small
    \setlength{\tabcolsep}{6pt}
    \centering
    \resizebox{\textwidth}{!}{
    \begin{tabular}{l|cc|cc|ccc|ccc}
        \toprule
        \multirow{2}{*}{Method} &
          \multirow{2}{*}{Bi-T} &
          \multirow{2}{*}{Gloss} &
          \multicolumn{2}{c|}{Input} &
          \multicolumn{3}{c|}{Phoenix-2014T (DTW$\downarrow$)} &
          \multicolumn{3}{c}{How2Sign (DTW$\downarrow$)} \\
        \cmidrule(lr){4-5} \cmidrule(lr){6-8} \cmidrule(lr){9-11}
                                   &              &          & Pose         & RGB      & Body          & Hand       & Avg           & Body       & Hand          & Avg        \\
        \midrule
        MotionGPT ~\cite{motiongpt} & \ding{51} & $\times$ & \ding{51} & $\times$ & 8.97          & 10.14      & 9.45          & 9.43       & 10.96         & 9.82       \\
        SOKE~\cite{soke} & $\times$ & $\times$ & \ding{51} & $\times$ & \textbf{2.58} & 5.89       & \textbf{4.26} & 7.92       & \textbf{3.07} & 5.49       \\
        T2S-GPT~\cite{t2sgpt} & $\times$ & $\times$ & \ding{51} & $\times$ & 7.32          & 9.86       & 8.28          & 7.15       & 11.21         & 8.49       \\
        MoMask~\cite{momask} & $\times$ & $\times$ & \ding{51} & $\times$ & 3.55          & 6.75       & 5.38          & -          & -             & -          \\
        NSA~\cite{nsa} & $\times$ & $\times$ & \ding{51} & $\times$ & \uline{3.09}   & 6.80       & 5.41          & 7.83       & 7.33          & 7.44       \\
        \midrule
        \rowcolor{aliceblue} SignGPT+TSA &
          \ding{51} &
          $\times$ &
          \ding{51} &
          $\times$ &
          3.31 &
          \textbf{4.98} &
          \uline{4.32} &
          \textbf{4.98} &
          \uline{4.11} &
          \textbf{4.76} \\
        \rowcolor{aliceblue} SignGPT (Ours)  & \ding{51} & $\times$ & \ding{51} & $\times$ & 3.35          & \uline{5.07} & 4.36          & \uline{5.05} & 4.18          & \uline{4.82} \\
        \bottomrule
    \end{tabular}
    }
    \label{tab:comparison-slg}
\end{table*}

\subparagraph{\textbf{Sign Language Translation.}}
The comparative SLT results are shown in Table~\ref{tab:comparison-slt}. On Phoenix-2014T, SignGPT trails SignLLM~\cite{signllm-slt} and MixSignGraph~\cite{mixsigngraph} on ROUGE-L and on development-set BLEU-4; on test BLEU-4, SignGPT+TSA slightly exceeds SignLLM but remains below MixSignGraph. Direct causal attribution is not possible because these systems differ in input modality, architectural priors, and training objectives. SignLLM incorporates a transmission-theory prior during codebook quantization; MixSignGraph uses RGB input, task-specific graph modules (LSG, TSG, and HSG), and Text-based CTC pre-training; SignGPT instead operates on pose input with a shared LLM backbone. These design differences, together with Phoenix-2014T's small and domain-specific setting, may contribute to the observed gaps and require controlled comparisons to disentangle.

In the selected examples in Fig.~\ref{fig:slt-slg}, the MotionGPT translations contain semantic mismatches, whereas the SignGPT translations remain closer to the reference text.

\begin{table*}[t]
    \caption{
        \textbf{Sign language translation (SLT) on Phoenix-2014T and How2Sign.} We report BLEU-4 (B4) and ROUGE-L (R). Bi-T indicates support for both sign-to-text translation and text-to-sign generation within one dataset-specific model. MotionGPT results are reproduced using our matched sign-language adaptation; all other baseline values are transcribed from the source studies cited for the corresponding results.
    }
    \small
    \centering
    \resizebox{\textwidth}{!}{
    \setlength{\tabcolsep}{6pt}
    \begin{tabular}{l|cc|cc|cccc|cc}
        \toprule
        \multirow{2}{*}{Method} &
          \multirow{2}{*}{Bi-T} &
          \multirow{2}{*}{Gloss} &
          \multicolumn{2}{c|}{Input} &
          \multicolumn{4}{c|}{Phoenix-2014T (Dev/Test)} &
          \multicolumn{2}{c}{How2Sign (Test)} \\
        \cmidrule(lr){4-5} \cmidrule(lr){6-9} \cmidrule(lr){10-11}
                                   &              &          & Pose         & RGB      & B4$\uparrow$ & R$\uparrow$ & B4$\uparrow$ & R$\uparrow$ & B4$\uparrow$ & R$\uparrow$ \\
        \midrule
        MotionGPT ~\cite{motiongpt}        & \ding{51} & $\times$ & \ding{51} & $\times$ & 11.53        & 27.14       & 10.98        & 27.05       & 8.72         & 28.61       \\
        SLTCC~\cite{sltcc}                      & $\times$     & $\times$ & \ding{51} & $\times$ & -            & -           & -            & -           & 11.8         & 31.1        \\
        SLT~\cite{slt}                        & $\times$     & \ding{51} & $\times$ & \ding{51} & 20.69    & 45.54 & 20.17      & \uline{45.34} & -         & -           \\
        CSGCR~\cite{csgcr}                      & $\times$     & $\times$ & $\times$     & \ding{51} & 15.08    & 38.96       & 15.18        & 38.85       & -            & -           \\
        Uni-Sign~\cite{uni-sign}                   & $\times$     & $\times$ & \ding{51} & \ding{51} & -        & -           & -            & -           & 14.9         & 36.0        \\
        SignLLM~\cite{signllm-slt}                    & $\times$     & $\times$ & $\times$     & \ding{51} & \textbf{25.25} & \uline{47.23} & 23.40 & 44.49 & - & -    \\
        MixSignGraph~\cite{mixsigngraph}                    & $\times$     & $\times$ & $\times$     & \ding{51} & \uline{24.87} & \textbf{51.71} & \textbf{24.02} & \textbf{51.14} & 10.41 & 28.01    \\
        \midrule
        \rowcolor{aliceblue} SignGPT+TSA &
          \ding{51} &
          $\times$ &
          \ding{51} &
          $\times$ &
          24.17 &
          42.96 &
          \uline{23.56} &
          42.13 &
          \textbf{16.77} &
          \textbf{38.25} \\
        \rowcolor{aliceblue} SignGPT (Ours) &
          \ding{51} &
          $\times$ &
          \ding{51} &
          $\times$ &
          23.58 &
          41.72 &
          22.95 &
          41.28 &
          \uline{16.42} &
          \uline{37.69} \\
        \bottomrule
    \end{tabular}%
    }
    \label{tab:comparison-slt}
\end{table*}

\subsection{Exploratory Evaluation of the One-Turn Sign-to-Sign Pipeline}
\label{sec:response-eval}

\subparagraph{\textbf{Evaluation Set and System Controls.}}
We select 1,000 non-duplicate question-form motion--text pairs from the How2Sign test split whose signed questions could plausibly serve as opening turns of a dialogue. ``Non-duplicate'' describes only the sampled test inputs; it does not claim that the question content is novel or previously unseen. We run both pipelines on every pair, yielding 1,000 generated responses per system. Each pipeline is given the pre-extracted ASL pose sequence as its input, while the paired English question is retained only as an evaluation reference. The sign model translates the pose sequence into English, the frozen LLaMA 3.2-1B-Instruct mediator stochastically generates a new English reply, and the sign model translates that reply into an ASL pose sequence. No reference response text or response motion is supplied. Consequently, all steps after the input question are zero-shot with respect to response-level pairs and targets, rather than zero-shot with respect to the pretrained SLT and SLG tasks themselves. For every metric requiring back-translation, we use a separately trained SignGPT checkpoint optimized exclusively for the SLT task on How2Sign and frozen before evaluation. This evaluator shares the dataset and architectural family with the evaluated SignGPT system; it receives no SLG or response-level supervision and is applied unchanged to both pipelines. We also hold the mediator checkpoint, response prompt, sampling configuration, and evaluator fixed across systems. These controls reduce variation from the non-sign components of the cascade, but the SignGPT-based evaluator is not architecture-independent and may retain distributional bias. We therefore interpret T2M, M2M, and LLM-AR only as diagnostic proxies. Given the absence of paired sign-to-sign response data, this setting asks a narrow question: can SLT and SLG capabilities learned without response-level supervision be composed into a one-turn response pipeline?

\subparagraph{\textbf{Stage-wise Diagnostics.}}
Owing to the absence of such data, the failure modes of a sign-to-sign pipeline cannot be localized by a single end-to-end number, so we instrument each of its three transformations separately. M2T is corpus-level BLEU-4 between the SLT output and the paired English reference question; it measures how faithfully the understanding stage preserves the query semantics that everything downstream depends on. T2T applies DialogRPT-updown~\cite{dialogue}, a learned human-preference ranker for dialogue turns, to the (translated question, generated reply) pair. With the mediator held fixed, T2T evaluates the resulting text-level question--reply pair; stochastic response generation means that it does not strictly isolate query-translation quality. T2M is cycle-consistency BLEU-4 between the English reply and the frozen evaluator's back-translation of the synthesized pose sequence; it measures how much reply semantics survives motion synthesis. Because the evaluator itself can introduce both false matches and false mismatches, T2M is a noisy diagnostic rather than an absolute measure of signing fidelity. M2M applies DialogRPT-updown to the (reference question, back-translated reply) pair, giving an end-to-end proxy in which the errors of all three stages compound. Finally, LLM-AR measures whether the back-translated reply remains semantically relevant to the reference question and can therefore credit valid lexical paraphrases that BLEU-based proxies may miss.

\subparagraph{\textbf{Participants, Recruitment, and Procedure.}}
We recruited 12 Deaf ASL users through online channels. Participants were 20--30 years old, and each reported at least 10 years of ASL use. Data collection was conducted remotely through our custom annotation system over a one-month window. Participants chose when to complete their sessions and were instructed to rate trials only when they felt sufficiently rested and attentive. The system assigned each of the 1,000 selected test-set question samples to exactly three different participants and balanced the allocation so that each participant received 250 samples. This produced 3,000 participant--sample assignments while retaining complete three-rater coverage of every sample. For each trial, the two rendered responses---one from MotionGPT and one from SignGPT---were shown with system identity concealed, and their presentation order was randomized.

\subparagraph{\textbf{Measures and Aggregation.}}
Participants rated each response separately using two single-item five-point scales (5 being the maximum). Motion smoothness targets only the temporal quality of the rendered signing: 1 indicates severe frame-to-frame jitter or visibly broken motion, and 5 indicates coherent motion with no perceptible jitter. Response appropriateness targets semantics rather than form: 1 indicates no recognizable relation to the question, and 5 indicates that the reply clearly and properly answers it; participants were instructed to ignore rendering artifacts for this item so that the two dimensions remain as separable as possible. Each assigned sample yielded four scalar ratings (two system responses $\times$ two rating items). Thus, each participant evaluated 500 response outputs and provided 1,000 scalar ratings; across the study, this yielded 6,000 participant--response observations and 12,000 scalar ratings. For each sample, system, and item, we first averaged the three participant ratings. The values in Table~\ref{tab:user-study} are unweighted descriptive means over the resulting 1,000 sample-level averages (equivalently, 3,000 raw ratings per system and item). Appendix~\ref{app:survey} documents the participant-facing instructions and rating items.

\begin{table}[!htbp]
\caption{Stage-wise automatic proxy metrics for the LLM-mediated single-turn ASL response pipeline on How2Sign. M2T compares the translated question against the reference question; T2T scores the (translated question, English reply) pair; T2M compares the English reply against the text back-translated by a frozen evaluator from its generated pose sequence; M2M scores the (reference question, back-translated reply) pair. B4 denotes BLEU-4 and DR denotes DialogRPT-updown. Both systems use the same frozen mediator, response prompt, sampling configuration, and frozen SLT evaluator.}
\label{tab:slc}
\centering
\small
\renewcommand{\arraystretch}{1.1}
\setlength{\tabcolsep}{6pt}
\begin{tabularx}{\linewidth}{l*{4}{>{\centering\arraybackslash}X}}
\toprule
\multirow{2}{*}{Method}
  & M2T & T2T & T2M & Pipeline M2M \\
\cmidrule(lr){2-5}
  & B4$\uparrow$ & DR$\uparrow$
  & B4$\uparrow$ & DR$\uparrow$ \\
\midrule
MotionGPT~\cite{motiongpt}
  & 10.26 & 0.681 & 4.90 & 0.089 \\
\rowcolor{aliceblue}
SignGPT (Ours)
  & \textbf{21.83} & \textbf{0.714}
  & \textbf{13.69} & \textbf{0.236} \\
\bottomrule
\end{tabularx}
\end{table}

\begin{table}[!htbp]
\caption{End-to-end response relevance and exploratory subjective ratings for the single-turn ASL response pipeline. For each system, LLM-AR is the percentage of its 1,000 outputs judged relevant by GPT-4o after the shared frozen SLT back-translation, with one binary judgment per output ($n=1{,}000$ per system). The subjective results are descriptive means from 12 Deaf participants. Every sample was rated by three different participants; each system--item mean therefore summarizes 1,000 three-rater sample averages (3,000 raw ratings). Response appropriateness examines whether the signed reply plausibly answers the signed question; motion smoothness examines whether the rendered motion is free of frame-to-frame jitter.}
\label{tab:user-study}
\centering
\small
\setlength{\tabcolsep}{6pt}
\renewcommand{\arraystretch}{1.3}

\begin{tabularx}{\linewidth}{l*{3}{>{\centering\arraybackslash}X}}
\toprule
\multirow{2}{*}{Method}
  & \shortstack{Response Relevance}
  & \multicolumn{2}{c}{Subjective Pilot (1--5)} \\
\cmidrule(lr){2-2}
\cmidrule(lr){3-4}
  & LLM-AR (\%)$\uparrow$
  & \shortstack{Response Appropriateness$\uparrow$}
  & \shortstack{Motion Smoothness$\uparrow$} \\
\midrule
MotionGPT~\cite{motiongpt}
  & 15.7
  & 2.71
  & 1.26 \\
\rowcolor{aliceblue}
SignGPT (Ours)
  & \textbf{52.2}
  & \textbf{3.67}
  & \textbf{4.35} \\
\bottomrule
\end{tabularx}
\end{table}

\subparagraph{\textbf{Results and Analysis.}}
Table~\ref{tab:slc} and Table~\ref{tab:user-study} report descriptive point estimates. The two systems have similar T2T scores under the shared mediator (0.714 vs.\ 0.681), whereas their end-to-end M2M proxy scores differ by a factor of 2.65 (0.236 vs.\ 0.089). The larger numerical separation appears after all stages are composed and is consistent with differences propagating through the cascade; it does not establish a particular scaling relationship between stage-wise and end-to-end scores.

In the relevance judgment, SignGPT produces a relevant reply for 52.2\% of its 1,000 outputs under the automated protocol. Because LLM-AR is measured after back-translation through an imperfect frozen evaluator, it conflates errors from the response pipeline and the evaluator; it is therefore a noisy proxy rather than a direct estimate or lower bound of human-perceived response effectiveness. Human raters assign SignGPT a mean response-appropriateness rating of 3.67 and a mean motion-smoothness rating of 4.35. This gap suggests that raters found temporal coherence stronger than semantic appropriateness under this protocol. The lower T2M score (13.69) than M2T score (21.83) is consistent with additional semantic loss during motion synthesis, although the two diagnostics operate at different stages and are not directly comparable measures of difficulty.

All estimates above are descriptive. A mean response appropriateness of 3.67 on a 5-point scale and an LLM-AR of 52.2\% both indicate that substantial failures remain even under our own evaluation protocol. We therefore position this study as an exploratory comparison of single-turn pipeline outputs and as a feasibility signal.

\begin{figure*}[!htbp]
\centering
\includegraphics[width=\textwidth]{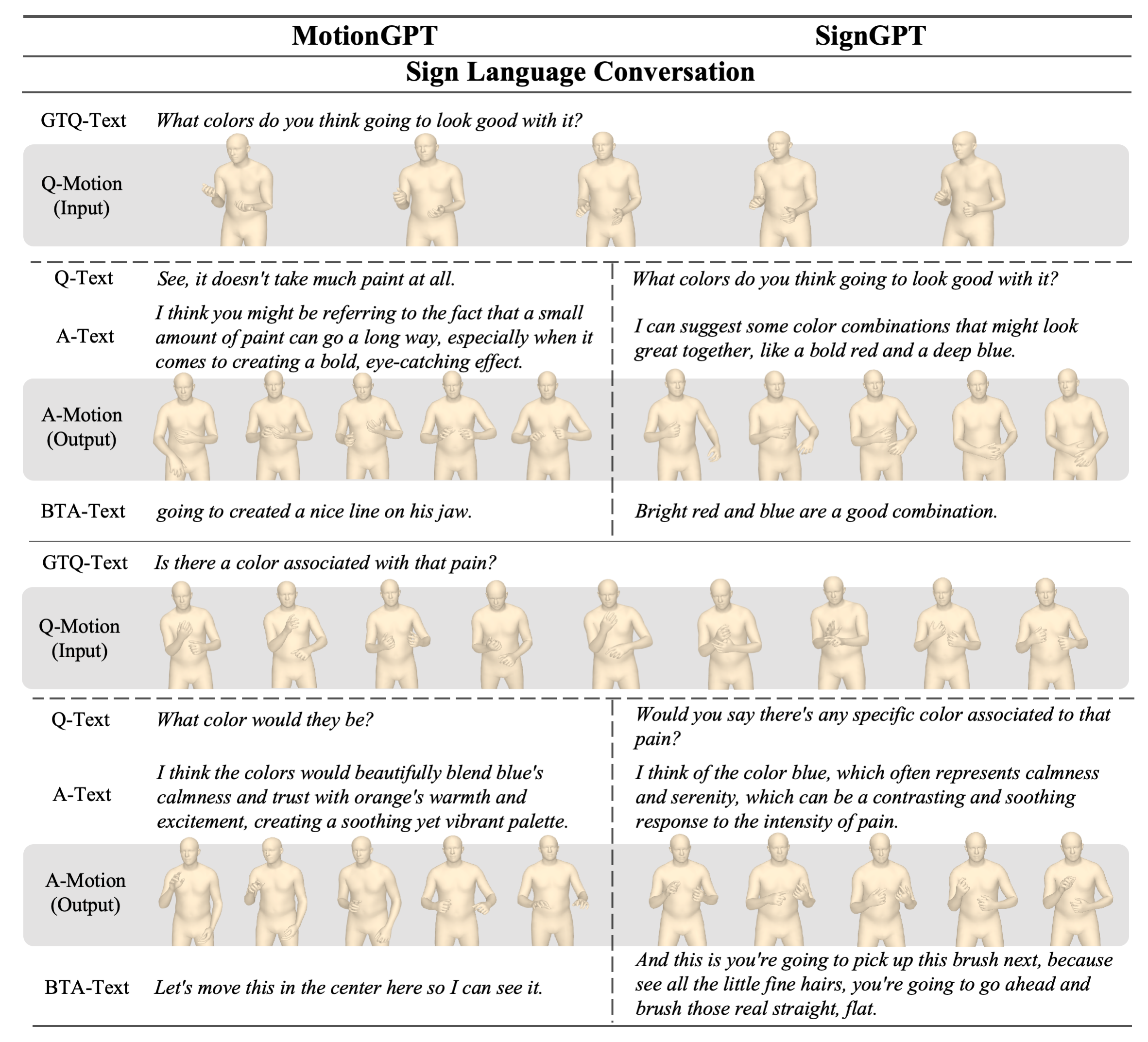}
\caption{Qualitative examples from the LLM-mediated, one-turn ASL response pipeline. GTQ-Text is the reference question; Q-Motion is the input query; Q-Text is the SLT output; A-Text is the response-model output; A-Motion is the SLG output; and BTA-Text is the frozen evaluator's back-translation. The examples show one semantically related response and one case exhibiting a discrepancy.}
\label{fig:slc}
\Description{Two one-turn response examples compare MotionGPT in the left column with SignGPT in the right column. Each example begins with a reference question and input signing frames, followed by the translated question, generated answer text, answer signing frames, and back-translated answer.}
\end{figure*}

\subsection{Ablation Studies}\label{sec:ablation}
\subparagraph{\textbf{PHVQ Components.}}
Table~\ref{tab:abl-phvq} reports cumulative PHVQ ablations. On Phoenix-2014T, where intermediate component variants are reported, part-aware codebooks improve all reconstruction metrics relative to the vanilla VQ-VAE, and the rows that subsequently add BM-TCN and the hand-specific losses show further gains. On How2Sign, the table compares only the vanilla VQ-VAE, the complete PHVQ tokenizer, and its TSA variant; it therefore does not isolate the individual components on that dataset. TSA leaves the reconstruction point estimates nearly unchanged on both datasets, consistent with its use as an optional embedding-alignment regularizer rather than a reconstruction component. Tables~\ref{tab:comparison-slg} and~\ref{tab:comparison-slt} separately report the downstream results of the complete variants. Appendix~\ref{app:additional-experiments} provides further ablations.

\begin{table*}[t]
\caption{\textbf{Ablation study on constructing PHVQ across two datasets.}
The blank row denotes the vanilla VQ-VAE baseline without any of the following components.
Part-Aware: modeling different body parts using independent codebooks (body and face, left hand, right hand);
TCN: temporal convolutional network for enhanced temporal modeling;
Hand-E: incorporating $\mathcal{L}_{\text{Pos}}$ and $\mathcal{L}_{\text{Angle}}$ to strengthen fine-grained quantization and reconstruction quality of fingers.
(Text-Space Alignment) TSA: adding the optional cosine loss between pooled motion projections and the frozen LLaMA input-embedding target during PHVQ training.}
\small
\centering
\resizebox{\textwidth}{!}{
\setlength{\tabcolsep}{6pt}
\begin{tabular}{l|cccc|cccc}
\toprule
\multirow{2}{*}{Dataset} &
\multicolumn{4}{c|}{Method} &
\multirow{2}{*}{MPJPE$\downarrow$} &
\multirow{2}{*}{PA-MPJPE$\downarrow$} &
\multirow{2}{*}{DTW-JPE$\downarrow$} &
\multirow{2}{*}{ACCEL$\downarrow$} \\
\cmidrule(lr){2-5}
& Part-Aware   & BM-TCN       & Hand-E       & TSA             &               &               &               &               \\
\midrule
\multirow{5}{*}{Phoenix-2014T}
& \multicolumn{4}{c|}{Vanilla VQ-VAE} & 45.5          & 30.9          & 4.52          & 39.4          \\
& \ding{51} &      &      &     & 26.7          & 19.3          & 2.42          & 35.2          \\
& \ding{51} & \ding{51} &      &     & 24.5          & 19.1          & 2.31          & 34.8          \\
& \cellcolor{aliceblue}\ding{51} & \cellcolor{aliceblue}\ding{51} & \cellcolor{aliceblue}\ding{51} & \cellcolor{aliceblue} & \cellcolor{aliceblue}\textbf{21.7} & \cellcolor{aliceblue}\textbf{16.9} & \cellcolor{aliceblue}\textbf{2.03} & \cellcolor{aliceblue}\uline{32.4}  \\
& \ding{51} & \ding{51} & \ding{51} & \ding{51} & \uline{22.0} & \uline{17.0}  & \uline{2.07}  & \textbf{32.3} \\
\midrule
\multirow{3}{*}{How2Sign}
& \multicolumn{4}{c|}{Vanilla VQ-VAE} & 43.1          & 28.6          & 4.45          & 8.82          \\
& \cellcolor{aliceblue}\ding{51} & \cellcolor{aliceblue}\ding{51} & \cellcolor{aliceblue}\ding{51} & \cellcolor{aliceblue} & \cellcolor{aliceblue}\textbf{18.3} & \cellcolor{aliceblue}\textbf{15.6} & \cellcolor{aliceblue}\textbf{1.66} & \cellcolor{aliceblue}\uline{6.17}  \\
& \ding{51} & \ding{51} & \ding{51} & \ding{51} & \uline{18.4} & \uline{15.9}  & \uline{1.68}  & \textbf{6.15} \\
\bottomrule
\end{tabular}%
}
\label{tab:abl-phvq}
\end{table*}

\begin{table*}[t]
\caption{\textbf{Cumulative SignGPT ablations on the Phoenix-2014T test set.} The first row is the adapted MotionGPT baseline. PHVQ+TSA uses paired sentence-level translations for text-space alignment during tokenizer training. GHMLM denotes joint LoRA-based SLT--SLG model adaptation, IFT denotes instruction fine-tuning, and Avg is the reported aggregate DTW-JPE.}
\label{tab:abl-signgpt}
\centering
\small
\setlength{\tabcolsep}{4pt}
\renewcommand{\arraystretch}{1.2}

\begin{tabularx}{\linewidth}{
  *{4}{>{\centering\arraybackslash}X}|
  *{3}{>{\centering\arraybackslash}X}
}
\toprule
\multicolumn{4}{c|}{Method}
  & SLG
  & \multicolumn{2}{c}{SLT} \\
\cmidrule(lr){1-4}
\cmidrule(lr){5-5}
\cmidrule(lr){6-7}
PHVQ
  & \shortstack{PHVQ+TSA}
  & GHMLM
  & \shortstack{GHMLM IFT}
  & Avg$\downarrow$
  & B4$\uparrow$
  & R$\uparrow$ \\
\midrule

\multicolumn{4}{c|}{MotionGPT}
  & 9.45
  & 10.98
  & 27.05 \\

\ding{51} & & &
  & 8.13
  & 12.54
  & 30.72 \\

\ding{51} & & \ding{51} &
  & 4.48
  & 22.18
  & 39.86 \\

\rowcolor{aliceblue}
\ding{51} & & \ding{51} & \ding{51}
  & \uline{4.36}
  & \uline{22.95}
  & \uline{41.28} \\

  & \ding{51} & \ding{51} & \ding{51}
  & \textbf{4.32}
  & \textbf{23.56}
  & \textbf{42.13} \\

\bottomrule
\end{tabularx}
\end{table*}

\subparagraph{\textbf{Effectiveness of SignGPT.}}
Table~\ref{tab:abl-signgpt} reports cumulative SignGPT ablations. Replacing the adapted MotionGPT tokenizer with PHVQ reduces Avg-DTW-JPE from 9.45 to 8.13 and increases BLEU-4 from 10.98 to 12.54. Adding GHMLM, which initializes motion embeddings from PHVQ-quantized features and uses AMTP to predict the three synchronized part streams, produces the largest subsequent changes (Avg-DTW-JPE 4.48; BLEU-4 22.18). Instruction fine-tuning yields further, smaller improvements (Avg-DTW-JPE 4.36; BLEU-4 22.95), and the TSA variant obtains the best point estimates.

\section{Limitations and Future Work}

\subparagraph{\textbf{Interactive Pipeline and User Evaluation.}}
Our current one-turn response pipeline provides an initial demonstration that independently learned sign language translation (SLT) and sign language generation (SLG) capabilities can be composed into a sign-to-sign response workflow without response-level supervision. The pipeline currently operates on pre-extracted poses and is intended to serve as a foundation for future interactive systems. Extending this framework to real-time communication will require the integration of raw-video perception and further investigation of practical interaction factors, including turn taking, clarification and repair mechanisms, and consistency across multiple turns. Another promising direction is to collect accurately annotated multi-party sign language conversation data and develop end-to-end models that jointly optimize sign language dialogue capabilities and their evaluation.

Our participant study provides complete three-rater coverage of all 1,000 evaluated samples and offers initial evidence regarding users' perceptions of response appropriateness and motion smoothness. The observed difference between these two dimensions further motivates the development of more fine-grained evaluation protocols that assess semantic accuracy, linguistic naturalness, temporal coordination, and motion quality as related but distinct aspects. Co-designing such protocols with Deaf participants will be particularly important to ensure that future evaluations reflect the priorities of Deaf communities and their real-world communication needs.

\subparagraph{\textbf{Responsible Use and Future Deployment.}}
Smooth and visually plausible motion does not necessarily guarantee semantic accuracy, particularly in high-stakes contexts such as medical, legal, educational, and emergency communication. The present results should therefore be viewed as a research-stage proof of concept rather than evidence of readiness for deployment in such settings. Interactive correction and clarification mechanisms could further improve system reliability and enhance user agency. In high-stakes applications, such systems should be designed to complement rather than replace qualified sign language interpreters, while preserving clear and direct access to professional interpreting services.

\section{Conclusion}
This paper introduced SignGPT, a unified pose-based framework that connects gloss-free sign-to-text translation and text-to-sign generation within a single model. Its Part-aware Hierarchical VQ-VAE (PHVQ) represents coordinated body, hand, and facial motion via body-to-hand hierarchical quantization and bidirectional multiscale temporal encoding, addressing the loss of fine-grained articulation and non-manual detail incurred by generic motion tokenizers. Its Gloss-free Heterogeneous Motion-aware Language Model (GHMLM) reuses PHVQ features directly as motion embeddings rather than learning newly initialized codebook embeddings, and Asymmetric Multi-Token Prediction (AMTP) allows a shared hidden state to be decoded into either text tokens or part-specific motion tokens. Experiments on How2Sign (ASL) and Phoenix-2014T (DGS) show that SignGPT attains competitive performance on the reported SLT and SLG metrics, and component ablations confirm the contribution of part-aware hierarchical quantization, temporal encoding, and feature reuse (RQ1, RQ2).

Coupling the two directions into an exploratory LLM-mediated sign-to-sign response pipeline, we find that raters judge the generated responses to be reasonably appropriate in content and reasonably smooth in motion, yet the ratings reveal that component-level accuracy does not translate directly into satisfactory interaction quality (RQ3): errors introduced during understanding propagate through text-mediated generation into synthesis, and subtle non-manual and prosodic cues remain difficult to render faithfully. We therefore view SignGPT less as a deployable system than as a shared testbed for studying error propagation across the sign-to-sign pipeline. Future work will extend evaluation to larger and more diverse samples of Deaf signers and community-grounded protocols, strengthen modeling of non-manual markers and discourse-level context, and move beyond text-mediated responses toward interaction designs that better preserve user agency.

\section{Acknowledgments}
The authors acknowledge the use of OpenAI's ChatGPT as an assistive tool for grammar checking, improving textual clarity, code development, and debugging. All AI-assisted text and code were reviewed and verified by the authors. In addition, GPT-4o was used as an automated evaluator in the exploratory sign-to-sign interaction study to assess the relevance of back-translated responses to their corresponding reference questions and to compute LLM-AR. AI tools were not involved in participant data collection or the analysis of participant feedback. The participant study complied with the ethics-review requirements applicable to the authors' research environment, and all participants provided informed consent.

\bibliographystyle{ACM-Reference-Format}
\bibliography{sample-base}

\appendix
\section{More Details of Evaluation Metrics}
To evaluate PHVQ reconstruction error, we use metrics common to motion capture and generation research~\cite{apdix1,apdix3,999,motiongpt}: Mean Per Joint Position Error (MPJPE), Procrustes-Aligned MPJPE (PA-MPJPE)~\cite{apdix2}, and acceleration error (ACCEL). MPJPE is the mean Euclidean distance between ground-truth and predicted joints after pelvis centering. PA-MPJPE applies a rigid alignment before computing this distance. ACCEL measures discrepancies in joint acceleration and serves as a proxy for temporal smoothness.

Given that the temporal length of generated sign language sequences often differs from that of the ground-truth data, we employ the well-established DTW algorithm \cite{apdix5} to account for these temporal misalignments when calculating joint position errors. This approach gives rise to the DTW-JPE (Dynamic Time Warping on Joint Position Errors) evaluation metric \cite{nsa,apdix6}. By finding an optimal temporal alignment path, DTW-JPE measures the sequence-level distance between the generated and ground-truth sign motions. It is used to assess the quality of sign language reconstruction tasks and to evaluate Sign Language Generation (SLG) performance.

\section{Additional Experiments}
\label{app:additional-experiments}
\subsection{Impact of Motion Representations.}

We begin from the H3D feature family used in motion-generation research~\cite{h3d}, whose full motion vector contains $(\dot{r}_a,\dot{r}_x,\dot{r}_z,r_y,\mathbf{j}_p,\mathbf{j}_v,\mathbf{j}_r,\mathbf{c}_f)$. Here, $\dot{r}_a$ is root angular velocity around the vertical axis; $\dot{r}_x$ and $\dot{r}_z$ are root linear velocities on the ground plane; $r_y$ is root height; $\mathbf{j}_p$ denotes root-relative joint positions; and $\mathbf{j}_v$, $\mathbf{j}_r$, and $\mathbf{c}_f$ denote per-joint velocities, rotations, and binary foot-contact states. We additionally use the 10-dimensional SMPL-X expression parameters $\mathbf{f}\in\mathbb{R}^{10}$. All four representation variants retain $(\dot{r}_a,\dot{r}_x,\dot{r}_z,r_y)$ and $\mathbf{f}$ and exclude $\mathbf{c}_f$. The ``Pos.,'' ``Rot.,'' and ``Vel.'' columns in Table~\ref{tab:input-format} denote the optional per-joint features $\mathbf{j}_p$, $\mathbf{j}_r$, and $\mathbf{j}_v$, respectively.

\begin{table*}[!htbp]
\caption{Ablation study on the impact of different motion representations on PHVQ reconstruction quality across two sign language datasets.}
\centering
\resizebox{\textwidth}{!}{
\renewcommand{\arraystretch}{1.1}
\setlength{\tabcolsep}{3pt}
\begin{tabular}{ccc|cccc|cccc}
\toprule
\multicolumn{3}{c|}{Input Format} & \multicolumn{4}{c|}{Phoenix-2014T} & \multicolumn{4}{c}{How2Sign} \\
\cmidrule(lr){1-3}\cmidrule(lr){4-7}\cmidrule(lr){8-11}
Pos. & Rot. & Vel. & MPJPE$\downarrow$ & PA-MPJPE$\downarrow$ & DTW-JPE$\downarrow$ & ACCEL$\downarrow$ & MPJPE$\downarrow$ & PA-MPJPE$\downarrow$ & DTW-JPE$\downarrow$ & ACCEL$\downarrow$ \\
\midrule
\ding{51} & \ding{51} & \ding{51} & \uline{23.1} & 18.2 & \uline{2.24} & \textbf{28.5} & \uline{22.3} & \uline{17.9} & \uline{1.97} & \uline{6.38} \\
\ding{51} &              & \ding{51} & 23.3 & \uline{17.9} & 2.26 & \uline{28.9} & 22.9 & 18.3 & 2.14 & 6.46 \\
\ding{51} & \ding{51} &              & 24.6 & 18.5 & 2.42 & 33.1 & 23.1 & 19.4 & 2.18 & 6.70 \\
\rowcolor{aliceblue} \ding{51} & & & \textbf{21.7} & \textbf{16.9} & \textbf{2.03} & 32.4 & \textbf{18.3} & \textbf{15.6} & \textbf{1.66} & \textbf{6.17} \\
\bottomrule
\end{tabular}%
}
\label{tab:input-format}
\end{table*}

Prior work~\cite{humantomato,151} observes that per-joint rotations, velocities, and foot-contact states can be derived from joint coordinates, making the hybrid representation partly redundant. We therefore ablate the per-joint representation components in Table~\ref{tab:input-format}. Across both datasets, the position-only variant gives the lowest MPJPE, PA-MPJPE, and DTW-JPE point estimates, while the richer representations yield mixed changes in ACCEL. We consequently use $\vec{\mathbf{m}}_i=(\dot{r}_a,\dot{r}_x,\dot{r}_z,r_y,\mathbf{j}_p,\mathbf{f})$, excluding per-joint velocities, per-joint rotations, and foot-contact states while retaining root angular velocity, ground-plane root linear velocity, and root height. With 72 non-root joints, this final representation has $1+2+1+216+10=230$ dimensions.

\subsection{Ablation Study of Codebook Size on PHVQ}

\begin{table*}[t]
\caption{PHVQ reconstruction results for different body--face and hand codebook sizes. We use $N_{BF}=128$ and $N_{LH}=N_{RH}=256$ in the remaining experiments.}
\label{tab:codebook}
\centering
\small
\renewcommand{\arraystretch}{1.1}
\setlength{\tabcolsep}{4pt}

\begin{tabularx}{\linewidth}{
  *{2}{>{\centering\arraybackslash}X}|
  *{3}{>{\centering\arraybackslash}X}|
  *{3}{>{\centering\arraybackslash}X}
}
\toprule
\multicolumn{2}{c|}{Codebook Size}
  & \multicolumn{3}{c|}{Phoenix-2014T (DTW$\downarrow$)}
  & \multicolumn{3}{c}{How2Sign (DTW$\downarrow$)} \\
\cmidrule(lr){1-2}
\cmidrule(lr){3-5}
\cmidrule(lr){6-8}
$N_{BF}$
  & $N_{LH}=N_{RH}$
  & Body & Hand & Avg
  & Body & Hand & Avg \\
\midrule
128 & 192
  & 1.89 & 2.55 & 2.26
  & 1.43 & 2.05 & 1.76 \\
128 & 320
  & 1.86 & 2.32 & \uline{2.14}
  & \uline{1.40} & \textbf{1.90} & \textbf{1.65} \\
160 & 256
  & \uline{1.85} & 2.49 & 2.18
  & 1.41 & 1.95 & 1.68 \\
96 & 256
  & 1.97 & \uline{2.41} & 2.23
  & 1.48 & 1.96 & 1.72 \\
\rowcolor{aliceblue}
128 & 256
  & \textbf{1.82} & \textbf{2.21} & \textbf{2.03}
  & \textbf{1.39} & \uline{1.93} & \uline{1.66} \\
\bottomrule
\end{tabularx}
\end{table*}

\begin{figure*}[t]
\centering
\includegraphics[width=\textwidth]{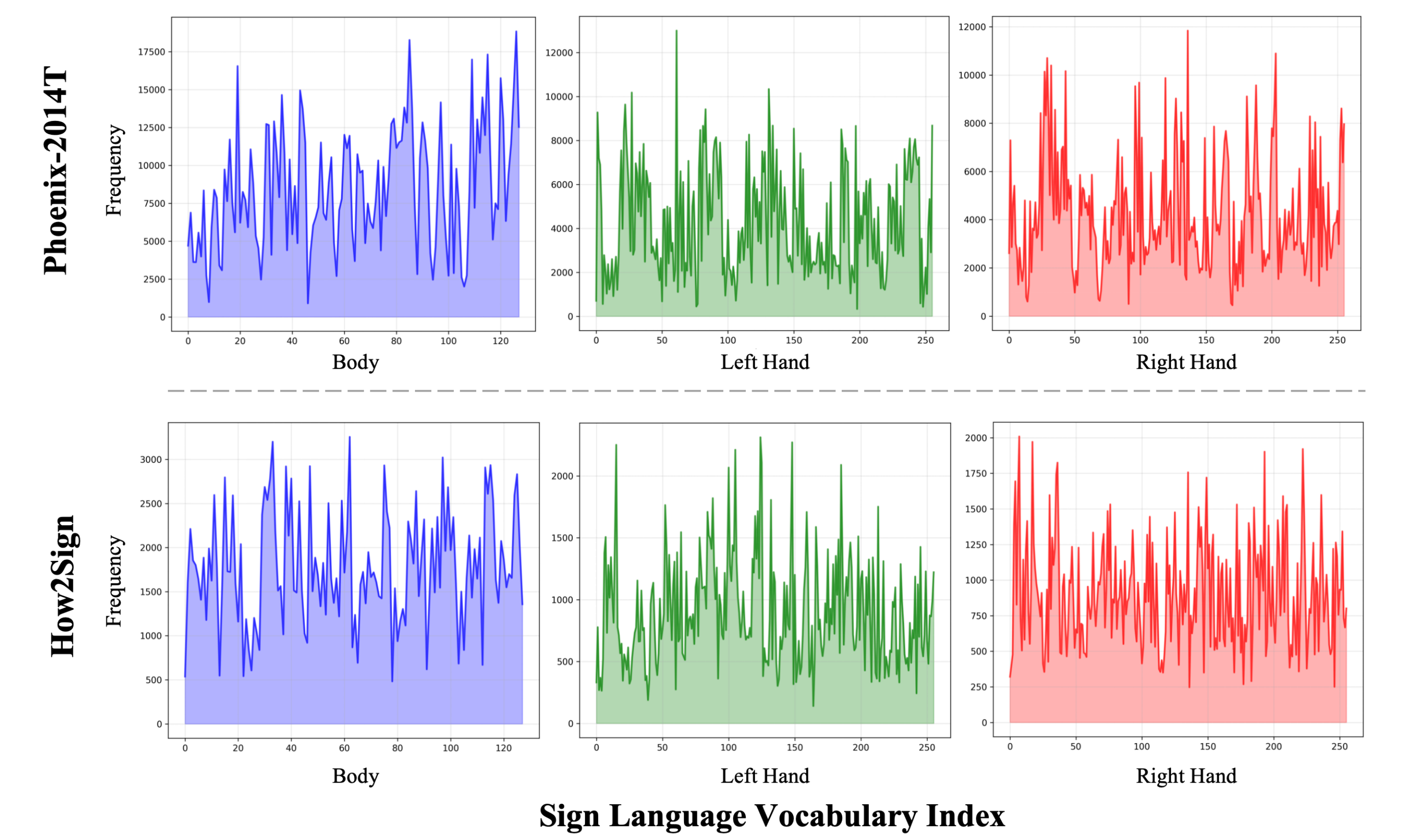}
\caption{Statistics of discrete tokens in the sign language motion vocabularies (codebooks) for different body parts after quantizing the training samples. For the Phoenix-2014T and How2Sign datasets, from left to right the represented body parts are body--face ($N_{BF}=128$), left hand ($N_{LH}=256$), and right hand ($N_{RH}=256$).}
\label{fig:codebook}
\Description{Six filled line plots show token frequency by codebook index. The top row covers Phoenix-2014T and the bottom row covers How2Sign; columns represent body-face, left-hand, and right-hand codebooks. Frequencies vary substantially across entries, but activity is distributed across the full index ranges rather than concentrated in only a few codes.}
\end{figure*}

We investigate how the codebook sizes for different body parts affect PHVQ reconstruction, as shown in Table~\ref{tab:codebook}. On Phoenix-2014T, $N_{BF}=128$ and $N_{LH}=N_{RH}=256$ gives the lowest point estimates for all three DTW measures among the tested settings. On How2Sign, $N_{BF}=128$ and $N_{LH}=N_{RH}=320$ gives the lowest hand and aggregate DTW, while the selected $128/256$ setting gives the lowest body DTW and a nearly identical aggregate value (1.66 vs.\ 1.65) with smaller hand output vocabularies. We therefore use $128/256$ across both datasets. This comparison supports the selected capacity but does not directly measure code redundancy or optimization difficulty.

In addition, we encode the training motions from Phoenix-2014T and How2Sign and count the occurrences of each discrete token, as shown in Fig.~\ref{fig:codebook}. Tokens occur across the full index ranges in all six plots, although their frequencies are non-uniform. This usage analysis describes code occupancy; it does not by itself establish the linguistic meaning or redundancy of individual codes.

\subsection{Effectiveness of BM-TCN}
We compare BM-TCN with two representative alternatives: (i) a plain 1D ResNet encoder following~\cite{t2mgpt}, which uses stacked fixed-kernel convolutions, and (ii) a causal TCN~\cite{131}, which restricts each temporal position to past frames. All variants have the same parameter count and are trained under identical settings. In Table~\ref{tab:tcn-ablation}, BM-TCN gives the lowest point estimate for every reported reconstruction metric on both datasets, the ResNet is second, and the causal TCN has the largest errors. This pattern is consistent with a benefit from combining bidirectional context with multiple temporal scales during offline tokenization. Because encoder directionality and receptive-field design differ together across these variants, the ablation does not establish that a particular linguistic cue causes the observed differences or that bidirectionality is necessary in every setting.

\begin{table*}[t]
\caption{Ablation study on different encoder designs in PHVQ.}
\label{tab:tcn-ablation}
\centering
\resizebox{\textwidth}{!}{
\renewcommand{\arraystretch}{1.0}
\setlength{\tabcolsep}{3pt}
\begin{tabular}{l|cccc|cccc}
\toprule
\multirow{2}{*}{Encoder} & \multicolumn{4}{c|}{Phoenix-2014T} & \multicolumn{4}{c}{How2Sign} \\
\cmidrule(lr){2-5}\cmidrule(lr){6-9}
 & MPJPE$\downarrow$ & PA-MPJPE$\downarrow$ & DTW-JPE$\downarrow$ & ACCEL$\downarrow$ & MPJPE$\downarrow$ & PA-MPJPE$\downarrow$ & DTW-JPE$\downarrow$ & ACCEL$\downarrow$ \\
\midrule
Causal TCN~\cite{131}           & 30.6 & 23.9 & 2.94 & 43.6 & 25.8 & 22.0 & 2.40 & 8.30 \\
 ResNet~\cite{t2mgpt}      & \uline{23.6} & \uline{17.4} & \uline{2.17} & \uline{33.9} & \uline{19.9} & \uline{16.1} & \uline{1.78} & \uline{6.45} \\
\rowcolor{aliceblue} BM-TCN (Ours) & \textbf{21.7} & \textbf{16.9} & \textbf{2.03} & \textbf{32.4} & \textbf{18.3} & \textbf{15.6} & \textbf{1.66} & \textbf{6.17} \\
\bottomrule
\end{tabular}
}
\end{table*}

\subsection{Hierarchical Quantization}
Another design choice in PHVQ is body-to-hand hierarchical quantization, which supplies quantized body--face features to the hand streams. We compare this direction with the hand-to-body strategy in HumanTOMATO~\cite{humantomato}. We also include an independent baseline in which the streams are quantized without cross-part conditioning~\cite{121,soke}. Table~\ref{tab:hier-direction} shows that body-to-hand conditioning gives the lowest point estimate for every reported metric on both datasets, while the independent variant is second. One possible explanation is that body and arm context is useful when representing fine-grained hand articulation; conversely, hand-to-body conditioning must transform two quantized hand streams together with the body stream. The current ablation, however, compares conditioning directions as complete variants and does not isolate information flow, noise propagation, or linguistic disambiguation as causal mechanisms.

\begin{table*}[t]
\caption{Ablation study on the conditioning direction of the hierarchical quantization in PHVQ.}
\label{tab:hier-direction}
\centering
\resizebox{\textwidth}{!}{
\renewcommand{\arraystretch}{1.0}
\setlength{\tabcolsep}{3pt}
\begin{tabular}{l|cccc|cccc}
\toprule
\multirow{2}{*}{Hierarchical Strategy} & \multicolumn{4}{c|}{Phoenix-2014T} & \multicolumn{4}{c}{How2Sign} \\
\cmidrule(lr){2-5}\cmidrule(lr){6-9}
 & MPJPE$\downarrow$ & PA-MPJPE$\downarrow$ & DTW-JPE$\downarrow$ & ACCEL$\downarrow$ & MPJPE$\downarrow$ & PA-MPJPE$\downarrow$ & DTW-JPE$\downarrow$ & ACCEL$\downarrow$ \\
\midrule
Hand-to-Body~\cite{humantomato}              & 23.2 & 17.5 & 2.14 & 33.0 & 19.6 & 16.2 & 1.75 & 6.30 \\
Independent~\cite{121,soke}                  & \uline{22.0} & \uline{17.3} & \uline{2.04} & \uline{32.7} & \uline{18.5} & \uline{16.0} & \uline{1.67} & \uline{6.22} \\
\rowcolor{aliceblue} Body-to-Hand (Ours)     & \textbf{21.7} & \textbf{16.9} & \textbf{2.03} & \textbf{32.4} & \textbf{18.3} & \textbf{15.6} & \textbf{1.66} & \textbf{6.17} \\
\bottomrule
\end{tabular}
}
\end{table*}

\subsection{Ablation Study of the Motion-Embedding Fusion Weight Beta}

When the language model requires motion embeddings as input, we use $\beta$ to weight the body--face and hand streams according to Equation~\ref{eq:fusion}. Table~\ref{tab:beta} compares three values. At $\beta=1/3$, all three streams receive equal per-stream weights; this setting gives the lowest Avg-DTW and the highest ROUGE-L on both datasets, as well as the highest How2Sign BLEU-4. Phoenix-2014T BLEU-4 is slightly higher at $\beta=0.4$ (23.07 vs.\ 22.95). We therefore select $\beta=1/3$ as the best aggregate trade-off among the tested settings. These single point estimates do not measure run-to-run stability, and the table does not by itself identify which linguistic information changes as $\beta$ varies.

\begin{table}[t]
\caption{Ablation study of SignGPT using different motion embedding fusion weights $\beta$ on two sign language datasets.}
\label{tab:beta}
\centering
\small
\setlength{\tabcolsep}{4pt}
\renewcommand{\arraystretch}{1.1}

\begin{tabularx}{\linewidth}{
  >{\centering\arraybackslash}X|
  >{\centering\arraybackslash}X|
  >{\centering\arraybackslash}X|
  *{2}{>{\centering\arraybackslash}X}
}
\toprule
\multirow{2}{*}{Dataset}
  & \multirow{2}{*}{$\beta$}
  & SLG
  & \multicolumn{2}{c}{SLT} \\
\cmidrule(lr){3-3}
\cmidrule(lr){4-5}
  & & Avg-DTW$\downarrow$ & B4$\uparrow$ & R$\uparrow$ \\
\midrule
\multirow{3}{*}{Phoenix-2014T}
  & 0.2 & 4.65 & 19.58 & 36.19 \\
  & \cellcolor{aliceblue}$1/3$
  & \cellcolor{aliceblue}\textbf{4.36}
  & \cellcolor{aliceblue}22.95
  & \cellcolor{aliceblue}\textbf{41.28} \\
  & 0.4 & 4.38 & \textbf{23.07} & 41.26 \\
\midrule
\multirow{3}{*}{How2Sign}
  & 0.2 & 5.32 & 13.81 & 35.27 \\
  & \cellcolor{aliceblue}$1/3$
  & \cellcolor{aliceblue}\textbf{4.82}
  & \cellcolor{aliceblue}\textbf{16.42}
  & \cellcolor{aliceblue}\textbf{37.69} \\
  & 0.4 & 4.91 & 16.20 & 37.35 \\
\bottomrule
\end{tabularx}
\end{table}

\subsection{Ablation study on motion embedding initialization}

A central design choice of GHMLM is to initialize motion embeddings from frozen PHVQ codebook features and the projectors $P^{\uparrow}_p$, rather than adding motion-token indices to the LLM vocabulary with randomly initialized embeddings~\cite{soke,sltcc}. We compare the two initialization schemes under the same architecture and training schedule. In \textbf{Random Init}, each motion token receives a randomly initialized embedding optimized by the language-modeling objective; in \textbf{Codebook Reuse}, its embedding is retrieved from the corresponding codebook $\mathcal{C}^{p}$ and projected into the LLM hidden space. Table~\ref{tab:emb-init} shows better SLG and SLT point estimates for Codebook Reuse on both datasets: BLEU-4 increases by 10.05 points on Phoenix-2014T and by 7.19 points on How2Sign, while Avg-DTW decreases from 10.54 to 4.36 and from 8.85 to 4.82, respectively. These results support tokenizer-informed initialization under the tested budget. They do not separately determine whether the gains arise from kinematic structure, optimization speed, or another difference between the initial embeddings.

\begin{table}[h]
\caption{Ablation study on the motion embedding initialization scheme of SignGPT. ``Random Init'' adds new motion tokens with randomly initialized embeddings; ``Codebook Reuse'' reuses frozen PHVQ codebook features projected by $P^{\uparrow}_p$.}
\label{tab:emb-init}
\centering
\small
\setlength{\tabcolsep}{4pt}
\renewcommand{\arraystretch}{1.1}

\begin{tabularx}{\linewidth}{
  >{\centering\arraybackslash}X|
  >{\centering\arraybackslash}X|
  >{\centering\arraybackslash}X|
  *{2}{>{\centering\arraybackslash}X}
}
\toprule
\multirow{2}{*}{Dataset}
  & \multirow{2}{*}{\shortstack{Init.\\Scheme}}
  & SLG
  & \multicolumn{2}{c}{SLT} \\
\cmidrule(lr){3-3}
\cmidrule(lr){4-5}
  & & Avg-DTW$\downarrow$ & B4$\uparrow$ & R$\uparrow$ \\
\midrule
\multirow{2}{*}{Phoenix-2014T}
  & Random Init
  & 10.54 & 12.90 & 26.10 \\
  & \cellcolor{aliceblue}Codebook
  & \cellcolor{aliceblue}\textbf{4.36}
  & \cellcolor{aliceblue}\textbf{22.95}
  & \cellcolor{aliceblue}\textbf{41.28} \\
\midrule
\multirow{2}{*}{How2Sign}
  & Random Init
  & 8.85 & 9.23 & 23.83 \\
  & \cellcolor{aliceblue}Codebook
  & \cellcolor{aliceblue}\textbf{4.82}
  & \cellcolor{aliceblue}\textbf{16.42}
  & \cellcolor{aliceblue}\textbf{37.69} \\
\bottomrule
\end{tabularx}
\end{table}

\subsection{Ablation study on body-part token decoding paradigms}
GHMLM employs Asymmetric Multi-Token Prediction (AMTP), in which the BF, LH, and RH tokens at each step share the same hidden state $h_k$ and are predicted in parallel by three heterogeneous heads. An alternative is to keep a single prediction head and decode the three body parts serially within each step (i.e., $\mathrm{BF} \!\rightarrow\! \mathrm{LH} \!\rightarrow\! \mathrm{RH}$), which triples the number of forward passes per frame but provides explicit conditional dependencies between parts. We therefore compare two configurations: (i) \textbf{Serial-AR}, a single head autoregressively producing BF, LH, RH in sequence within every step; and (ii) \textbf{AMTP} (ours), which fuses the three part embeddings via $\mathcal{F}(\cdot;\beta)$ before the LLM forward pass and predicts all parts in parallel from the shared $h_k$.

As shown in Table~\ref{tab:parallel-vs-serial}, AMTP gives lower Avg-DTW and higher BLEU-4 and ROUGE-L than Serial-AR on both datasets. By construction, AMTP uses one shared backbone evaluation per motion step, whereas Serial-AR uses three sequential evaluations. The comparison supports parallel multi-head decoding under the tested configuration, but it does not isolate whether the metric differences arise from prediction order, separate heads, fused context, or the different computation budgets.

\begin{table}[h]
\caption{Comparison between parallel multi-head decoding (AMTP) and serial single-head autoregressive decoding of body-part motion tokens.}
\label{tab:parallel-vs-serial}
\centering
\small
\setlength{\tabcolsep}{4pt}
\renewcommand{\arraystretch}{1.1}

\begin{tabularx}{\linewidth}{
  >{\centering\arraybackslash}X|
  >{\centering\arraybackslash}X|
  >{\centering\arraybackslash}X|
  *{2}{>{\centering\arraybackslash}X}
}
\toprule
\multirow{2}{*}{Dataset}
  & \multirow{2}{*}{\shortstack{Decoding\\Scheme}}
  & SLG
  & \multicolumn{2}{c}{SLT} \\
\cmidrule(lr){3-3}
\cmidrule(lr){4-5}
  & & Avg-DTW$\downarrow$ & B4$\uparrow$ & R$\uparrow$ \\
\midrule
\multirow{2}{*}{Phoenix-2014T}
  & Serial-AR
  & 5.48 & 14.59 & 28.00 \\
  & \cellcolor{aliceblue}AMTP (ours)
  & \cellcolor{aliceblue}\textbf{4.36}
  & \cellcolor{aliceblue}\textbf{22.95}
  & \cellcolor{aliceblue}\textbf{41.28} \\
\midrule
\multirow{2}{*}{How2Sign}
  & Serial-AR
  & 5.71 & 10.44 & 25.56 \\
  & \cellcolor{aliceblue}AMTP (ours)
  & \cellcolor{aliceblue}\textbf{4.82}
  & \cellcolor{aliceblue}\textbf{16.42}
  & \cellcolor{aliceblue}\textbf{37.69} \\
\bottomrule
\end{tabularx}
\end{table}

\section{Implementation Details}
\label{app:implementation}
\subsection{PHVQ Hierarchical Quantization Strategy}
\label{app:phvq-quantization}
Each PHVQ stream uses one codebook lookup; there are no residual quantization levels. Thus, the hierarchy described below is the conditioning order across body--face and hand streams, not an RVQ stack.
We first quantize $Z_{\text{BF}}$ using $\text{Q}_{\text{BF}}(\cdot)$ and codebook $\mathcal{C}_{\text{BF}}$ to obtain $\hat{Z}_{\text{BF}}$. The de-quantized body--face features are passed through $\text{Transform}(\cdot)$ and concatenated with the original left- and right-hand latent representations, $Z_{\text{LH}}$ and $Z_{\text{RH}}$. Dedicated convolutional layers produce the two fused hand representations, which are quantized once using $\text{Q}_{\text{LH}}(\cdot)$ and $\text{Q}_{\text{RH}}(\cdot)$ with codebooks $\mathcal{C}_{\text{LH}}$ and $\mathcal{C}_{\text{RH}}$. For each stream $p$, the quantized feature follows the reconstruction path $\bar{Z}_p=P^{\downarrow}_p(P^{\uparrow}_p(\hat{Z}_p))$. The three $\bar{Z}_p$ representations are concatenated and projected through $\text{Conv}_{1\times1}(\cdot)$ into $Z_{\text{unified}}$, which is passed to $\mathcal{D}_T$ to reconstruct $\widetilde{\mathcal{M}}$. Together, these operations form $\mathcal{R}_T$. The optional text-space cosine loss attaches to the output of $P^{\uparrow}_p$; the projector--reconstructor path itself is present in both SignGPT and SignGPT+TSA.

\subsection{PHVQ Loss Function}
\label{app:phvq-loss}
\subparagraph{\textbf{Joint Angle Loss.}}
Standard reconstruction loss typically calculates L1 distance based on joint Cartesian coordinates (XYZ positions). While this ensures overall pose accuracy, it is not sufficiently sensitive to unnatural states such as local bone rotations and joint hyperextension. Particularly in regions with complex joint structures like hands, minor positional errors can lead to visually unnatural finger bone bending. \(\mathcal{L}_{\text{Angle}}\) directly constrains joint angles, thereby constraining the anatomical correctness of generated gestures to some extent and improving the generation quality of fine hand details. We predefine 30 finger joint triplets \((p, j, c)\) covering all fingers of both hands, representing the parent joint, current joint, and child joint respectively. For each triplet, we calculate the angle formed by vectors \(\vec{v}_1=P_p-P_j\) and \(\vec{v}_2=P_c-P_j\). Given predicted joint positions \(J_{\text{pred}}\) and reference joint positions \(J_{\text{ref}}\), we compute the predicted angle \(\theta_{\text{pred}}\) and reference angle \(\theta_{\text{ref}}\) for each triplet. To ensure numerical stability, the angle is obtained using the \(\text{atan2}\) function based on the cross product and dot product of the normalized directional vectors \(\hat{v}_1 = \frac{\vec{v}_1}{\|\vec{v}_1\|}\) and \(\hat{v}_2 = \frac{\vec{v}_2}{\|\vec{v}_2\|}\):
\[
\theta = \text{atan2}\left(\|\hat{v}_1 \times \hat{v}_2\|, \hat{v}_1 \cdot \hat{v}_2\right)
\]
The final joint angle loss is obtained by calculating the Smooth L1 Loss between the predicted angle sequence and the reference angle sequence.
\[
\mathcal{L}_{\text{Angle}} = \text{Smooth L1}\left(\{\theta_{\text{pred}}\}, \{\theta_{\text{ref}}\}\right)
\]

\subparagraph{\textbf{Optional Text-Space Alignment Loss.}}
This optional regularizer uses no separate text encoder. We use the frozen input-embedding layer of the same LLaMA 3.2-1B backbone employed by GHMLM. Each paired sentence is tokenized in its original corpus language---German for Phoenix-2014T and English for How2Sign---without machine translation or cross-dataset language normalization. If $E(t_{ij})$ is the embedding of token $j$ in sample $i$ and $a_{ij}$ is its non-padding mask, the global text target is $T_{\mathrm{global}}^{(i)}=(\sum_j a_{ij})^{-1}\sum_j a_{ij}E(t_{ij})$. Part-specific projectors $P^{\uparrow}_p$ for $p\in\{\mathrm{BF},\mathrm{LH},\mathrm{RH}\}$ map the quantized latent representations into the common embedding space $\mathbb{R}^{d_s}$, with $d_s=d_{\mathrm{LLM}}=2048$.

For each stream $p$, we mean-pool $P^{\uparrow}_p(\hat{Z}_p)$ over its valid temporal positions to obtain $M_{\mathrm{global},p}^{(i)}$. We then average the complementary cosine similarity over samples and the three streams:

\[
\mathcal{L}_{\text{Cos}} = \frac{1}{3B} \sum_{i=1}^{B}\sum_{p\in\{\mathrm{BF},\mathrm{LH},\mathrm{RH}\}} \left( 1 - \frac{M_{\text{global},p}^{(i)} \cdot T_{\text{global}}^{(i)}}{\|M_{\text{global},p}^{(i)}\|_2 \|T_{\text{global}}^{(i)}\|_2} \right)
\]

Once the pooled representations are available, computing the three cosine terms costs $O(Bd_s)$ and does not use negative sampling. Each $P^{\downarrow}_p$ maps the projected representation back into its codebook space for the reconstruction path. The base SignGPT model sets the cosine-loss weight to zero; SignGPT+TSA enables it.

\begin{figure*}[!htbp]
\centering
\includegraphics[width=\textwidth]{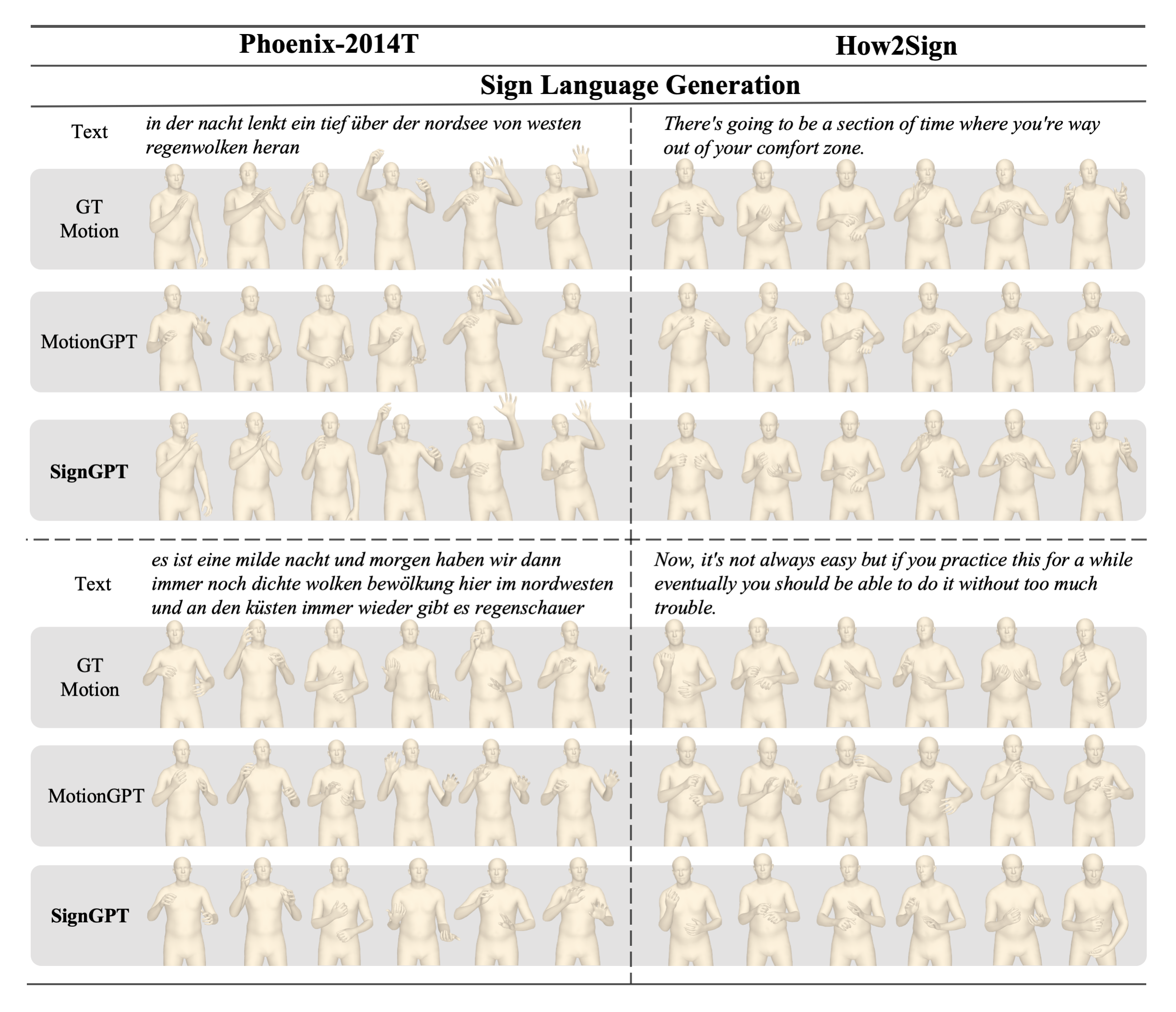}
\caption{Qualitative comparison between SignGPT and MotionGPT for sign language generation on the test sets of Phoenix-2014T (left) and How2Sign (right).}
\label{fig:more_slg}
\Description{Two generation examples for each dataset compare ground-truth signing frames with MotionGPT and SignGPT outputs. For both German weather text and English instructional text, SignGPT reproduces hand positions and upper-body trajectories that more closely follow the reference sequences, whereas MotionGPT exhibits larger pose and hand-shape deviations.}
\end{figure*}

\begin{figure*}[!htbp]
\centering
\includegraphics[width=\textwidth]{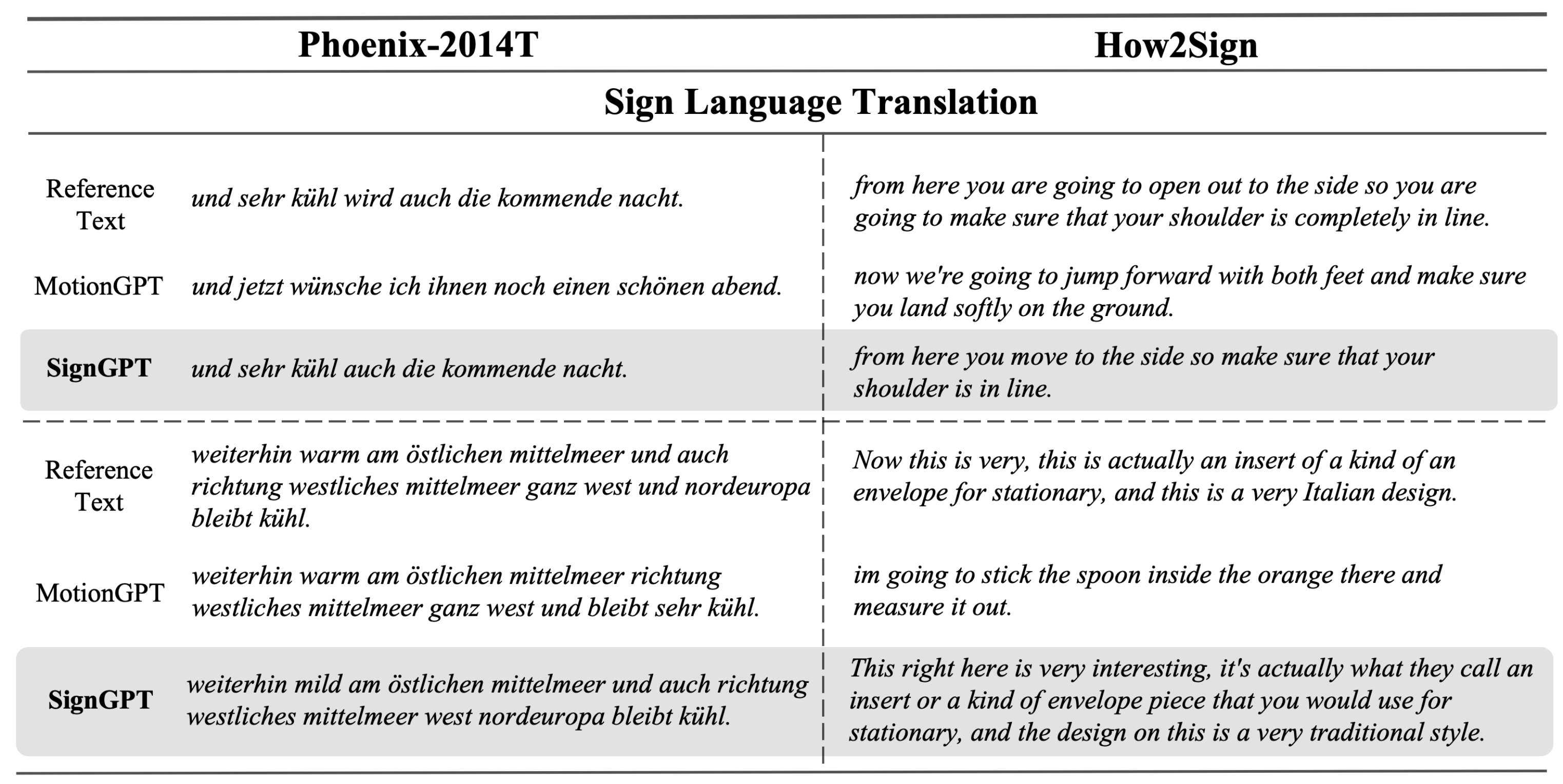}
\caption{Qualitative comparison between SignGPT and MotionGPT for sign-to-text translation on the test sets of Phoenix-2014T (left) and How2Sign (right).}
\label{fig:more_slt}
\Description{Two translation examples per dataset list the reference sentence followed by MotionGPT and SignGPT outputs. The SignGPT sentences retain more of the reference meaning in both German weather forecasts and English instructional descriptions, while MotionGPT introduces unrelated or contradictory content.}
\end{figure*}

\subsection{Model Configuration}
For Part-aware Hierarchical VQ-VAE (PHVQ), the number of Bidirectional Multi-scale Temporal Convolutional Network (BM-TCN) layers in both the encoder and decoder is set to 3. Specifically, each encoder begins with an initial $1{\times}3$ projection convolution that lifts the part-specific input ($110$-dim for the body-face stream and $60$-dim for each hand stream) to a $512$-dim hidden representation, followed by two stacked downsampling stages. Each downsampling stage consists of a strided $1{\times}3$ convolution (stride $=2$) and a BM-TCN module of depth $3$ with exponentially growing dilation rates $\{1, 3, 9\}$ (i.e., $r = 3$), where every dilated layer adopts symmetric padding and is wrapped with weight normalization, ReLU activation, and dropout. A final $1{\times}3$ convolution then projects the features to a $1024$-dim latent, yielding an output of shape $(B, 1024, T/4)$ with a temporal downsampling rate of $l = 4$. The decoder is fully mirror-symmetric to the encoder, replacing strided convolutions with nearest-neighbor upsampling while keeping the same BM-TCN depth and dilation schedule, and finally reconstructing the original $230$-dim full-body motion features at frame resolution $(B, 230, T)$. Owing to the exponential dilation schedule, each BM-TCN module attains a receptive field of $53$ frames, and the cascaded two-stage hierarchy further enlarges the effective receptive field to several hundred frames, sufficient to cover phrase-level temporal context in sign language. We set $\lambda_1=0.02$, $\lambda_3=0.5$, and $\beta=1/3$; $\lambda_2=0$ for SignGPT and $\lambda_2=0.02$ for SignGPT+TSA. During tokenizer training, the window size is set to $64$. For the Phoenix-2014T and How2Sign datasets, the input motion representations have a maximum length of $196$ and a minimum length of $24$. PHVQ has 54.5M parameters, and the inference complexity for a single sample (batch=1, input motion length=64) is 5.5 GFLOPs. The training times on Phoenix-2014T and How2Sign are 2.5 hours and 8 hours, respectively.

GHMLM has 1,250M parameters; for single-sample inference (batch=1, maximum sequence length=192), the computational complexity is 580.357 GFLOPs. The expanded token embedding size (vocabulary size) is 128,896. SignGPT (GHMLM) requires a total of 13 hours and 33 hours for LoRA tuning and instruction tuning on Phoenix-2014T and How2Sign, respectively.

\section{More Qualitative Results for SLT and SLG}
We present additional qualitative visualization results of SignGPT on the SLG and SLT tasks in Fig.~\ref{fig:more_slg} and Fig.~\ref{fig:more_slt}, respectively.

\section{Survey Instrument for the Exploratory Subjective Rating Study}
\label{app:survey}

This appendix documents the recruitment and allocation procedure,
participant-facing instructions, and item-level response scales for the
exploratory subjective rating study described in
Section~\ref{sec:response-eval}.

\subsection{Participants, Recruitment, and Allocation}

We recruited 12 Deaf ASL users online (age 20--30; at least 10 years of ASL
use). During the one-month remote study, participants completed sessions only
when sufficiently rested and attentive. The annotation system assigned each of
1,000 selected test-set question samples to exactly three different participants, balanced
at 250 samples per participant. Each sample contained one response per system;
system labels were concealed and response order was randomized per trial.

\subsection{Study Instructions}

Participants received the following instructions:

\begin{quote}
In each trial, you will view a signed question (together with its corresponding translation) and two rendered signed responses. The systems that produced these responses will not be identified. Rate
each response separately using both questions below. The questions assess
different aspects of the response. For response appropriateness, focus on
whether the response plausibly answers the signed question and, as far as
possible, ignore rendering artifacts. For motion smoothness, focus only on the
temporal continuity of the rendered motion and ignore whether the response is
semantically correct.
\end{quote}

\subsection{Item-Level Rating Questions}

For each rendered response, participants answered the following two questions.

\begin{enumerate}
    \item \textbf{Response appropriateness.} How appropriately does this signed
    response answer or otherwise plausibly respond to the signed question?
    Focus on the meaning of the response and ignore rendering artifacts.
    \begin{itemize}
        \item 1: No recognizable relation to the question.
        \item 2: Mostly unrelated or does not answer the question.
        \item 3: Partially related or ambiguous as a response.
        \item 4: Clearly related and mostly answers the question.
        \item 5: Clearly and properly answers or responds to the question.
    \end{itemize}

    \item \textbf{Motion smoothness.} How smooth is the motion in this rendered
    signed response? Focus on frame-to-frame continuity and transitions, and
    ignore whether the response answers the question.
    \begin{itemize}
        \item 1: Severe frame-to-frame jitter or visibly broken motion.
        \item 2: Frequent jitter or disrupted transitions.
        \item 3: Some noticeable jitter, but generally continuous motion.
        \item 4: Mostly smooth motion with only minor artifacts.
        \item 5: Coherent motion with no perceptible jitter.
    \end{itemize}
\end{enumerate}

No preference question was used; both responses were rated separately on both
items, producing four scalar ratings per trial. Each participant completed 250
trials (500 response outputs; 1,000 scalar ratings), for totals of 3,000 trials,
6,000 response observations, and 12,000 scalar ratings. For each system and
item, Table~\ref{tab:user-study} reports the mean over 1,000 sample-level
averages, each formed from three raters (equivalent to 3,000 raw ratings).

\end{document}